\pdfoutput=1

\documentclass[11pt]{article}

\usepackage[]{acl}

\usepackage{times}
\usepackage{tabularx}
\usepackage{latexsym}
\usepackage{amsmath}
\usepackage{amsfonts}
\usepackage{hyperref}
\usepackage{multirow}
\usepackage{graphicx}
\usepackage{makecell} 
\usepackage{adjustbox}
\usepackage{colortbl}
\usepackage{booktabs}
\usepackage{listings}
\usepackage{enumitem}
\usepackage[T1]{fontenc}
\usepackage[utf8]{inputenc}
\usepackage{ltablex}
\usepackage{microtype}
\usepackage{float}
\usepackage{longtable}
\usepackage{efbox,graphicx}
\usepackage{arydshln}
\usepackage{pifont}
\newcommand{\cmark}{\ding{51}}%
\efboxsetup{linecolor=gray,linewidth=1pt}
\newcommand{\xmark}{\ding{55}}%
\newcommand\cincludegraphics[2][]{\raisebox{-0.3\height}{\includegraphics[#1]{#2}}}

\title{Visualizing the Obvious: A Concreteness-based Ensemble Model \\ for Noun Property Prediction}

\author{Yue Yang\thanks{~~Equal Contribution.}~~, Artemis Panagopoulou$^{*}$, \\ \textbf{Marianna Apidianaki}, \textbf{Mark Yatskar}, \textbf{Chris Callison-Burch}\\
Department of Computer and Information Science, University of Pennsylvania\\
{\small \tt {\{yueyang1, artemisp, marapi, ccb, myatskar\}@seas.upenn.edu}}}

\begin{document}
\maketitle
\begin{abstract}
Neural language models encode rich knowledge about entities and their relationships which can be extracted from their representations using probing. Common properties of nouns (e.g., {\it red strawberries}, {\it small ant}) are, however, more challenging to extract compared to other types of knowledge because they are rarely explicitly stated in texts.
We hypothesize this to mainly be the case for perceptual properties which are obvious to the participants in the communication. We propose to extract these properties from images and use them in an ensemble model, in order to complement the information that is extracted from 
language models. We consider perceptual properties to be more concrete than abstract properties (e.g., {\it interesting}, {\it flawless}). We propose to  use the adjectives' concreteness score  as a lever to calibrate the contribution of each source (text vs. images). We evaluate our ensemble model in a ranking task where the actual properties of a noun need to be  ranked higher than other non-relevant properties. 
Our results show that the proposed combination of text and images greatly improves noun property prediction compared to powerful text-based language models.\footnotemark 
\end{abstract}
\section{Introduction}

\begin{figure}[!th]
\centering
    \includegraphics[width=7.7cm]{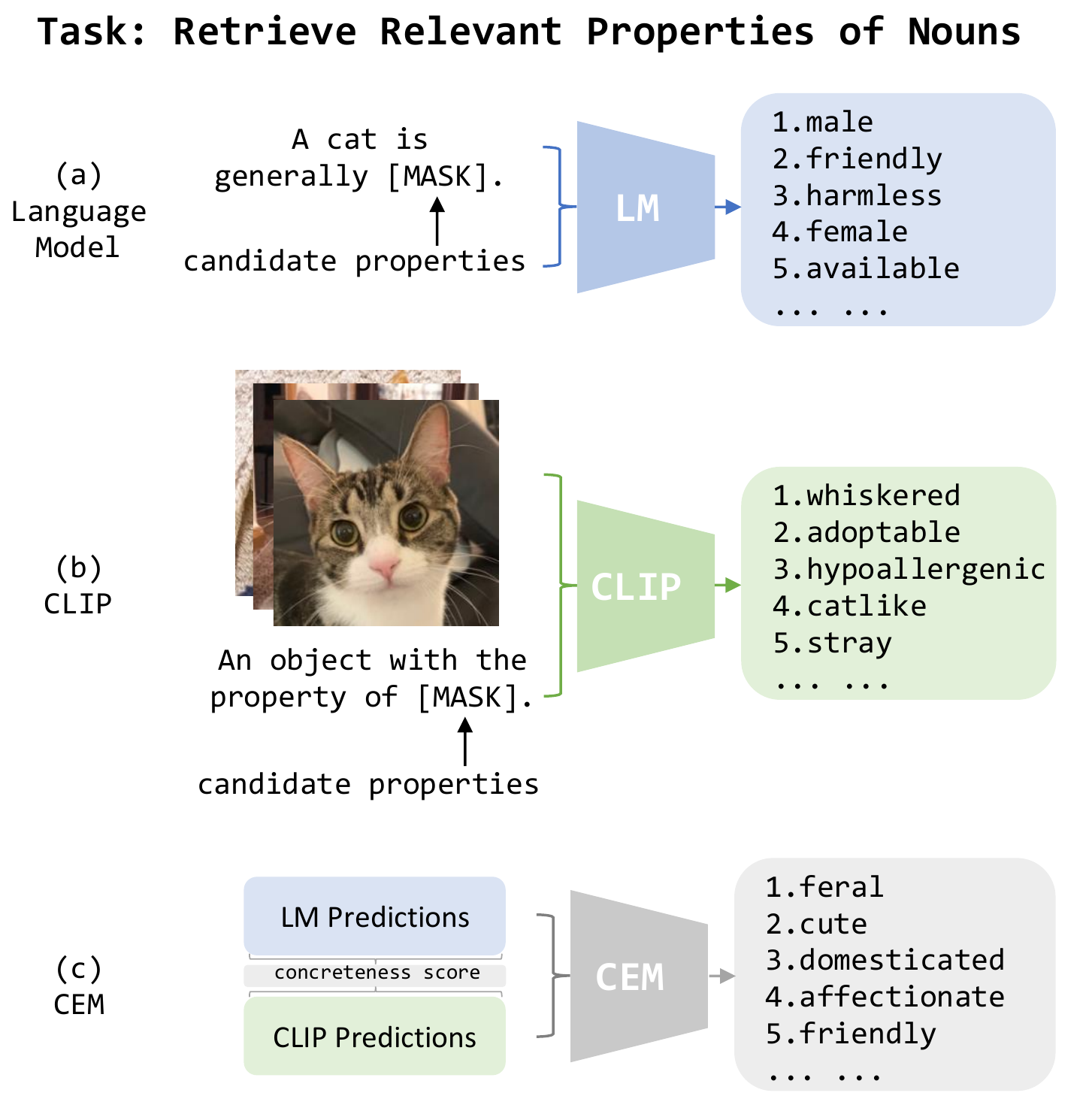}
    \caption{Our task is to retrieve relevant properties of nouns from a set of candidates. We tackle the task using (a) Cloze-task probing; (b)  {\sc CLIP} to compute the similarity between the properties and images of the noun; (c) a Concreteness Ensemble Model ({\sc CEM}) to ensemble language and {\sc CLIP} predictions  which relies on properties'  concreteness ratings. \protect}
    \label{fig:pipeline}
    \vspace{-.15cm}
\end{figure}

Common properties of concepts or entities (e.g., ``{\it These strawberries are {\underline {red}}}'') are rarely explicitly stated in texts, contrary to more specific properties which bring new information in the communication (e.g., ``{\it These strawberries are {\underline {delicious}}}''). This phenomenon, known as ``reporting bias'' \cite{GordonandVanDurme:2013,shwartz-choi-2020-neural}, makes it difficult to learn, or retrieve, perceptual properties from text. 
However, noun property identification is an important task which may allow AI applications to perform commonsense reasoning in a way that matches people's psychological or cognitive predispositions and can improve agent communication \citep{lazaridou-etal-2016-red}. Furthermore, identifying noun properties can contribute to better modeling concepts and entities, learning affordances (i.e. defining the possible uses of an object based on its qualities or properties), and understanding models' knowledge about the world.
~\footnotetext{Code and data are available at \url{https://github.com/artemisp/semantic-norms}}
Models that combine different modalities provide a sort of grounding which helps to alleviate the reporting bias problem \cite{kiela-etal-2014-improving,lazaridou-etal-2015-combining,zhang2022visual}. For example, multimodal models are better at predicting color  attributes compared to text-based language models \cite{paik-etal-2021-world,norlund-etal-2021-transferring}. Furthermore, visual representations of concrete objects improve performance in downstream NLP tasks  \cite{hewitt-etal-2018-learning}. Inspired by this line of work, we  
expect  concrete visual properties of nouns to be more accessible through images, and text-based language models to better encode abstract semantic properties.
We propose an ensemble model which combines information from these two sources for English noun property prediction.

We frame property identification as a ranking task, where relevant properties for a noun need to be retrieved from a set of candidate properties found in association norm datasets  \cite{mcrae2005semantic,devereux2014centre,norlund-etal-2021-transferring}.
We experiment with text-based language models \cite{devlin-etal-2019-bert,radford2019language,liu2019roberta}  and with {\sc CLIP} \cite{radford2021learning} which we query  using a slot filling task, as shown in Figures \ref{fig:pipeline}(a) and (b). Our ensemble model (Figure \ref{fig:pipeline}(c)) combines the strengths of the language and vision models, by specifically privileging the former or latter type of representation depending on the concreteness of the processed properties \cite{brysbaert2014concreteness}. Given that concrete properties are characterized by a higher degree of imageability  \cite{friendly1982toronto}, our model trusts the visual model for perceptual and highly concrete properties (e.g., color adjectives: {\it red}, {\it green}), and the language model for abstract properties (e.g., {\it free}, {\it infinite}). 
Our results confirm that {\sc CLIP} can identify nouns'  perceptual properties better than language models, which contain higher-quality information about abstract properties. 
Our ensemble model, which combines the two sources of knowledge, outperforms the individual models on the property ranking task by a significant margin. 

\section{Related Work}

Probing has been widely used in previous work for exploring the semantic knowledge that is encoded in language models. 
A common approach has been to convert  the facts, properties, and relations found in external knowledge sources into  ``fill-in-the-blank'' cloze statements, and to use them to query   language models. \citet{apidianaki-gari-soler-2021-dolphins} do so for nouns' semantic properties and  highlight how challenging it is to retrieve  this kind of information from {\sc BERT} representations   \cite{devlin-etal-2019-bert}. 

Furthermore, slightly different prompts tend 
to retrieve different semantic information \citep{ettinger-2020-bert}, compromising the robustness of semantic probing tasks. 
We propose to mitigate these problems by also relying on images.

Features extracted from different modalities can complement the information found in texts. Multimodal distributional models, for example, 
have been shown to outperform text-based approaches 
on semantic benchmarks \cite{silberer-etal-2013-models, Brunietal:2014,lazaridou-etal-2015-combining}. 
Similarly, ensemble models that integrate  multimodal and  text-based models outperform models that only rely on one modality in tasks such as visual question answering \cite{tsimpoukelli2021multimodal,alayrac2022flamingo,yang2021visual}, visual entailment \cite{song2022clip}, reading comprehension, natural language inference \cite{zhang2021semi,kiros2018illustrative}, text generation \cite{su2022language}, word sense disambiguation \cite{barnard2005word}, and video retrieval \cite{yang2021induce}.

We extend this investigation to noun property prediction.  
We propose a novel noun property retrieval model which combines information from language and vision models, and tunes their respective contributions based on property concreteness 
\cite{brysbaert2014concreteness}. 

Concreteness is a graded notion that  strongly correlates with the degree of imageability~\cite{friendly1982toronto,byrne1974item};  
concrete words generally tend to refer to tangible objects 
that the senses can easily perceive \cite{paivio1968concreteness}. 
We extend this idea to noun properties and hypothesize that vision models would have better knowledge of 
perceptual, and more concrete, properties (e.g., {\it red}, 
{\it flat}, {\it round}) than text-based language models, which would better 
capture 
abstract properties (e.g., {\it free}, {\it inspiring}, {\it promising}). 
We evaluate our ensemble model using concreteness scores 
  automatically predicted by a  regression model \cite{charbonnier2019predicting}. We compare these results to the  performance of the ensemble model with manual (gold) concreteness ratings  \cite{brysbaert2014concreteness}. 
In previous work, concreteness was measured based on the idea that abstract concepts relate to varied and composite situations \citep{Barsalou2004SituatingAC}. Consequently,  visually grounded representations of abstract concepts (e.g., {\it freedom}) should be more complex and diverse than those of concrete words (e.g., {\it dog})  \cite{lazaridou-etal-2015-combining,kiela-etal-2014-improving}.  
 
\citet{lazaridou-etal-2015-combining} 
specifically measure the entropy of the vectors induced by multimodal models which serve as an expression of how varied the information they encode is. They demonstrate that the entropy of multimodal vectors strongly correlates with the degree of abstractness of words.

\section{Experimental Setup}
\subsection{Task Formulation}
\label{taskformulation}

Given a noun $\mathcal{N}$ and a set of candidate properties $\mathbb{P}$, a model needs to select the 
properties $\mathbb{P}_{\mathcal{N}} \subseteq \mathbb{P}$  
that apply to  
$\mathcal{N}$. 
The candidate properties are the set of all adjectives retained from a resource (cf. Section \ref{datasetssection}), which characterize different nouns. A model needs to rank properties that apply to $\mathcal{N}$ higher than properties that apply to other nouns in the  resource. We consider that a property correctly characterizes a noun, if this property has been proposed for that noun by the annotators. 

\subsection{Datasets}
\label{datasetssection}

{\sc Feature Norms}: The \citet{mcrae2005semantic} dataset contains feature norms for 541 objects annotated by 725 participants. We follow \citet{apidianaki-gari-soler-2021-dolphins} and only use the {\sc is}\_{\sc adj} features of noun concepts, 
where the adjective describes a noun property. In total, there are 509 noun concepts with at least one {\sc is}\_{\sc adj} feature, and 209 unique properties. The {\sc Feature Norms} dataset  contains both perceptual properties (e.g., {\it tall}, {\it fluffy}) and non-perceptual ones (e.g., {\it intelligent}, {\it expensive}).
\medbreak
\noindent {\sc Memory Colors}: The dataset contains 109 nouns with an associated image and its corresponding prototypical color. There are 11 colors in total. 
\cite{norlund-etal-2021-transferring}. The data were 
scraped from existing knowledge bases on the web.
\medbreak
\noindent {\sc Concept Properties}: 
This dataset was created at the Centre for Speech, Language and  Brain \cite{devereux2014centre}. It contains concept property norm annotations collected from 30 participants. The data comprise 601 nouns with 400 unique properties. We keep aside 50 nouns (which are not in {\sc Feature Norms} and {\sc Memory Colors}) as our development set ({\tt dev}).   
 We use the {\tt dev} for prompt selection 
and hyper-parameter tuning. 
We call the rest of the dataset {\sc Concept Properties}-{\tt test} and use it for evaluation. 
\medbreak
\noindent {\sc Concreteness Dataset}: 
The \citet{brysbaert2014concreteness} dataset contains manual concreteness ratings for 37,058 English word lemmas and 2,896 two-word expressions, gathered through crowdsourcing. The original concreteness scores range from 0 to 5. We map them to $[0, 1]$ by dividing each score by 5. 

\subsection{Models}
\subsubsection{Language Models (LMs)}
We query language models about their knowledge of noun properties using cloze-style prompts (cf. Appendix \ref{lm prompts}). These contain the nouns in singular or plural form, and the \texttt{[MASK]} token at the position where the property should appear (e.g., ``{\it Strawberries are} \texttt{[MASK]}''). 
A language model assigns a probability score to a candidate property by relying on the wordpieces preceding and following the \texttt{[MASK]} token, $\textbf{W}_{\setminus t} = (w_1, ..., w_{t-1}, w_{t+1}, ..., w_{|\textbf{W}|})$:\footnote{We also experiment with the Unidirectional Language Model (ULM) which yields the probability of the masked token conditioned on the past tokens $\textbf{W}_{<t} = (w_1, ..., w_{t-1})$}
\begin{equation}
    \text{Score}_\text{LM}(\mathcal{P}) = \log P_{\text{LM}} (w_t = \mathcal{P} | \textbf{W}_{\setminus t}),
\end{equation}
where $P_{\text{LM}}(\cdot)$ is the probability from language model. We experiment with {\sc BERT-large}~\cite{devlin-etal-2019-bert}, {\sc RoBERta-large}~\cite{liu2019roberta}, {\sc GPT2-large}~\cite{radford2019language} and {\sc GPT3-davinci}, which have been shown to deliver impressive performance in Natural Language Understanding tasks~\cite{yamada2020luke,takase2021lessons,aghajanyan2021muppet}. 

\begin{table}[!t]
\centering
\resizebox{7.7cm}{!}{%
\begin{tabular}{ccccc}
\Xhline{3\arrayrulewidth}
\textbf{Dataset} & \textbf{\# $\mathcal{N}$s} & \textbf{\# $\mathcal{P}$s} & \textbf{$\mathcal{N}$-$\mathcal{P}$ pairs} & \textbf{$\mathcal{P}$s per $\mathcal{N}$} \\ \hline

{\sc Feature Norms}     &  509  &  209    &  1592  &  3.1  \\
{\sc Concept Properties}   &  601  &  400    &  3983  &  6.6 \\
{\sc Memory Colors} & 109 & 11 & 109 & 1.0
 \\ \Xhline{3\arrayrulewidth}
\end{tabular}
}
\caption{Statistics of the ground-truth datasets. We show the number of nouns (\# $\mathcal{N}$s), properties (\# $\mathcal{P}$s) and noun-property pairs ($\mathcal{N}$-$\mathcal{P}$ pairs), as well as the average number of properties per noun in each dataset.
}
\label{table: dataset stat}
\end{table}

Our property ranking setup allows to consider multi-piece adjectives (properties)\footnote{{\sc BERT}-type models split some words into multiple word pieces during tokenization (e.g., {\it colorful} $\rightarrow$ [`{\it color}',`{\it ful}']) \cite{wu2016googles}.} which were excluded from open-vocabulary masking experiments \cite{petroni-etal-2019-language,Bouraoui2020,apidianaki-gari-soler-2021-dolphins}. 
Since the candidate properties are known, we can obtain a score for a property composed of $k$ pieces ($\mathcal{P} = (w_t, ..., w_{t+k})$, $k \geq 1$) by taking the average of the scores assigned by the LM  to each piece:
\begin{equation}
\label{eq:avg_score}
    \text{Score}_\text{LM}(\mathcal{P}) = \frac{1}{k} \sum_{i=0}^{k} \log P_{\text{LM}} (w_{t+i} | \textbf{W}_{\setminus t + i})
\end{equation}

\noindent We report the results in Appendix \ref{app: multitok_performance} and show that our model is better than  other models at retrieving multi-piece properties. 

\subsubsection{Multimodal Language Models (MLMs)}

\par{\textbf{Vision Encoder-Decoder}} MLMs are 
language models conditioned on other modalities than text, for example, images. 
For each noun $\mathcal{N}$ in our datasets, we collect a set of 
images $\mathbb{I}$ from the web.\footnote{More details about the image collection procedure are given in 
Section \ref{implementation}.} 
We probe an MLM 
similarly to LMs, 
using the same set of prompts. 
An MLM yields a score for 
each property given an image $i \in \mathbb{I}$ using Formula \ref{eq:mlm_score}.
\begin{equation}
\label{eq:mlm_score}
    \text{Score}_\text{MLM}(\mathcal{P}, i) =  \log P_{\text{MLM}} (w_t = \mathcal{P} | \textbf{W}_{\setminus t}, i), 
\end{equation}
where $P_{\text{MLM}}(\cdot)$ is the probability from multimodal language model. In addition to the context $\textbf{W}_{\setminus t}$, the MLM conditions on the image $i$.\footnote{We use the same averaging method shown in equation \ref{eq:avg_score} to handle multi-piece adjectives for MLMs.} Then we aggregate over all the images $\mathbb{I}$ for  the noun $\mathcal{N}$ 
to get the score for the property:
\begin{equation}
    \text{Score}_\text{MLM}(\mathcal{P}) = \frac{1}{|\mathbb{I}|} \sum_{i \in \mathbb{I}} \text{Score}_\text{MLM}(\mathcal{P}, i)
\end{equation}

\noindent {\textbf{ViLT}} We experiment with the Transformer-based \cite{vaswani2017attention} {\sc ViLT} model~\cite{kim2021vilt} as an MLM.{\sc ViLT} uses the same tokenizer as {\sc BERT} and is pretrained on the Google Conceptual Captions (GCC) dataset which contains more than 3 million image-caption pairs for  about 50k words \cite{sharma2018conceptual}. Most other vision-language datasets contain a significantly smaller vocabulary (10k words).\footnote{The vocabulary size is much smaller than in {\sc BERT}-like models which are trained on a minimum of 8M words.} In addition, {\sc ViLT} 
requires minimal image pre-processing and is an open visual vocabulary model.\footnote{Open visual vocabulary 
models  do not need elaborate image pre-processing via an image detection pipeline. As such, they are not restricted to the object classes that are recognized by the pre-processing pipeline.} This contrasts with other multimodal architectures which require visual predictions before passing the images on to the multimodal layers \cite{li2019visualbert, lu2019vilbert, tan2019lxmert}. These have been shown to only marginally surpass text-only models \cite{yun-etal-2021-vision-language}.

\begin{figure}[!t]
\centering
    \includegraphics[width=7.7cm]{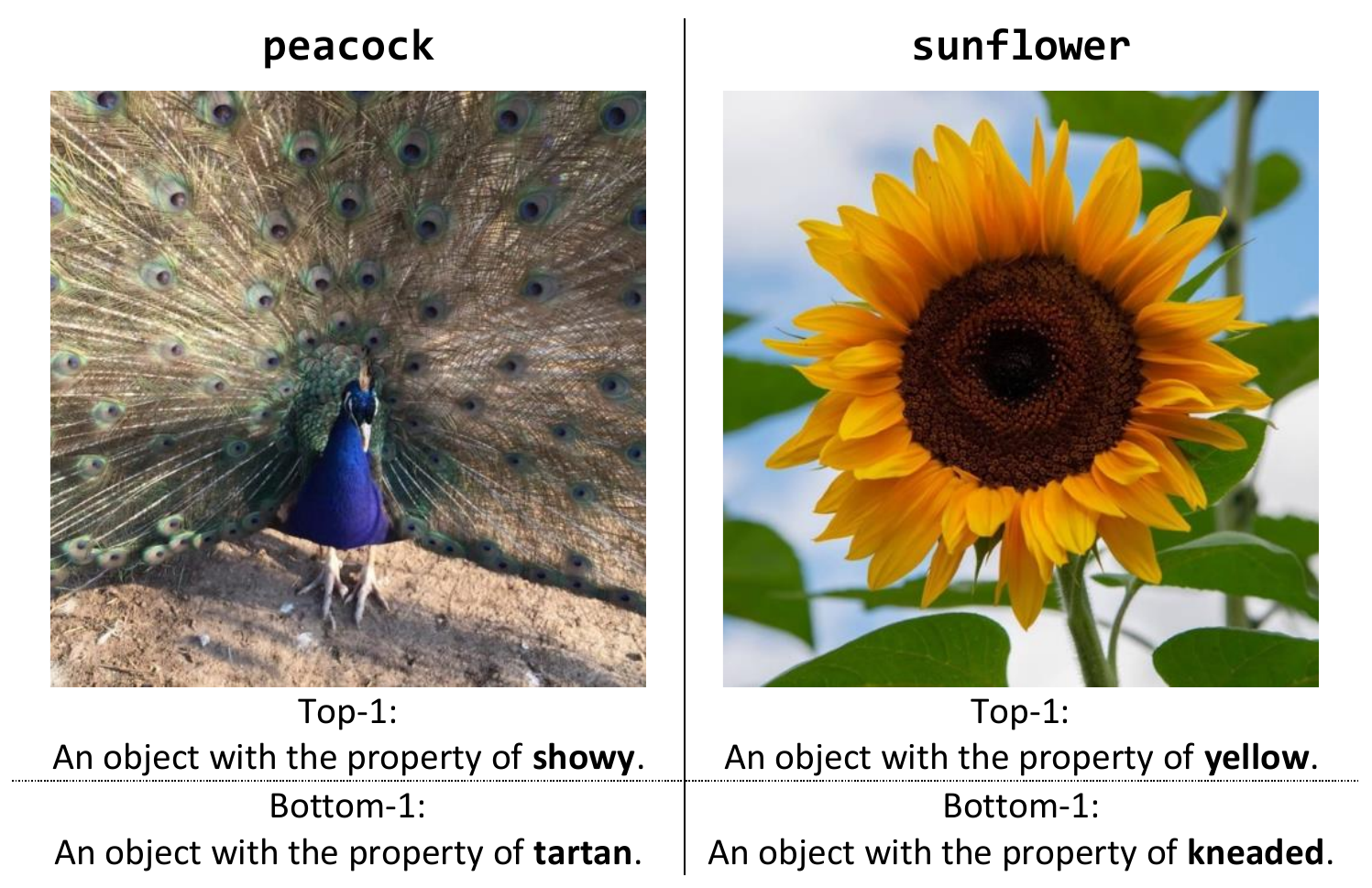}
    \caption{Examples of Top-1 and Bottom-1 prompts ranked by {\sc CLIP}.}
    \label{fig:clip_example}
\end{figure}
\medbreak
\noindent {\textbf{CLIP}} We also use the {\sc CLIP} model which is pretrained on 400M image-caption pairs  \cite{radford2021learning}. {\sc CLIP} is trained to align the embedding spaces learned from images and text using contrastive loss as a learning objective. The {\sc CLIP} model integrates a text encoder $f_{\text{T}}$ and a visual encoder $f_{\text{V}}$ which separately encode the text and image to vectors with the same dimension. Given a batch of image-text pairs, {\sc CLIP} maximizes the cosine similarity for matched pairs while minimizing the cosine similarity for unmatched pairs.

 We use {\sc CLIP} to compute the cosine similarity of an image $i \in \mathbb{I}$ and this text prompt ($s_\mathcal{P}$): ``\textit{An object with the property of} \texttt{[MASK]}'',  where the \texttt{[MASK]} token is replaced with a  candidate 
 property $\mathcal{P} \in \mathbb{P}$. 
The score for each property $\mathcal{P}$ is the mean similarity between the sentence prompt $s_\mathcal{P}$ and all images $\mathbb{I}$ collected for a noun:
\begin{equation}
    \text{Score}_\text{CLIP}(\mathcal{P}) = \frac{1}{|\mathbb{I}|} \sum_{i \in \mathbb{I}} \text{cos}(f_{\text{T}}(s_\mathcal{P}), f_{\text{V}}(i)) 
\end{equation}
This score serves to rank the candidate properties according to their relevance for a specific noun. Figure \ref{fig:clip_example} shows the most and least relevant properties for the nouns {\it peacock} and {\it sunflower}. 

\subsubsection{Concreteness Ensemble Model ({\sc CEM})}
\label{thecemmodel}

The concreteness score for a property guides {\sc CEM}  towards ``trusting'' 
the language or the vision model more. We propose two {\sc CEM} flavors 
which we describe as {\sc CEM-Pred} and {\sc CEM-Gold}. 
{\sc CEM-Pred} uses the score ($c_\mathcal{P} \in [0, 1]$) that is proposed by our  concreteness prediction model  
for every candidate property $\mathcal{P} \in \mathbb{P}$, 
while {\sc CEM-Gold} uses the score for $\mathcal{P}$ 
in the
\citet{brysbaert2014concreteness} dataset.\footnote{Properties in {\sc Memory Colors} have the highest average concreteness scores (0.82), followed by properties  in {\sc Feature Norms} (0.64) and {\sc Concept Properties} (0.62).} If there is no gold score for a property, we use the score of the word with the longest matching subsequence in the dataset.\footnote{This heuristic only applies to 15 (out of 209) properties in the  {\sc Feature Norms} dataset, and to 49 (out of 400) properties in {\sc Concept Properties}-{\tt test}. All 11 properties in {\sc Memory Colors} have a gold concreteness value.}  The idea behind this heuristic is that properties without  ground truth concreteness scores often have inflected forms or derivations in the dataset  (e.g.,  \textit{sharpened/sharpen},  \textit{invented/invention}, etc.).\footnote{It might happen that this heuristic matches antonymous words. Note that although these words have different meaning, they often have similar concreteness values (e.g., ``happy": 2.56, ``unhappy": 2.04; ``moral": 1.69, ``immoral": 1.59).} We also experimented with {\sc GloVe} word embedding cosine similarity which resulted in suboptimal performance (cf. Section \ref{sec:evaluation}). Additionally, sequence matching is much faster than {\sc GloVe} similarity (cf. Appendix \ref{app:inference_times}).

Both {\sc CEM}s combine the rank\footnote{The rank of a property $\mathcal{P}$ with respect to a model $\mathcal{M}$ denoted as $\text{Rank}_{\mathcal{M}}(\mathcal{P})$ is defined as the index of property $P$ in the list of all properties $\mathbb{P}$ sorted by decreasing $\text{Score}_\mathcal{M}(\mathcal{P})$.} of 
$\mathcal{P}$ 
proposed by the language model ($\text{Rank}_{\text{LM}}$) and by {\sc CLIP} ($\text{Rank}_{\text{CLIP}}$) through a weighted sum which is controlled by the concreteness score, $c_\mathcal{P}$:
\begin{equation} \label{rankequation}
    \begin{split}
        \text{Rank}_{\text{CEM}}(\mathcal{P}) = (1& - c_\mathcal{P})\cdot\text{Rank}_{\text{LM}} (\mathcal{P}) \\
        &+ c_\mathcal{P} \cdot \text{Rank}_{\text{CLIP}}(\mathcal{P})
    \end{split}
\end{equation}

\subsubsection{Concreteness Prediction Model} 

We generate concreteness scores using the model of \citet{charbonnier2019predicting} with FastText embeddings
\cite{bojanowski-etal-2017-enriching}. 
\begin{table}[!t]
\centering
\resizebox{7.7cm}{!}{%
\begin{tabular}{cc}
\Xhline{3\arrayrulewidth}
\textbf{Model}   & \textbf{Prompt Selected}  \\ \hline
{\sc BERT}    & Most \texttt{[NOUN-plural]} 
are \texttt{[MASK]}. \\
{\sc RoBERTa} & A/An \texttt{[NOUN-singular]} 
is generally \texttt{[MASK]}. \\
{\sc GPT-2}   & Most \texttt{[NOUN-plural]} 
are \texttt{[MASK]}. \\
{\sc ViLT}    & \texttt{[NOUN-plural]} 
are \texttt{[MASK]}. \\
{\sc CLIP}    & An object with the property of \texttt{[MASK]}. \\ \Xhline{3\arrayrulewidth}
\end{tabular}
}
\caption{The prompt template selected for each model.}
\label{table: final prompt selection}
\end{table}
The model leverages part-of-speech and suffix features to predict concreteness in a classical regression setting. We train the model on the 40k concreteness dataset \cite{brysbaert2014concreteness}, excluding the 425 adjectives found in our test sets. The model obtains a high Spearman $\rho$ correlation of 0.76 with the ground truth scores of the adjectives in our test sets. This result shows that automatically predicted scores are a viable alternative which allows the application of the method to new data and domains where hand-crafted resources might be unavailable.

\subsubsection{Baselines}
We compare the predictions of the language, vision, and ensemble models to the predictions  of three baseline methods.  
\medbreak
\noindent\textbf{{\sc Random}:} Generates a { \sc Random } property ranking  for each noun. 
\medbreak
\noindent\textbf{{\sc GloVE}:} Ranking based on the cosine similarity of the { \sc GloVE } embeddings \cite{pennington2014glove} of the noun and the property.
\medbreak
\noindent \textbf{{\sc Google Ngram:}} Ranking by the bigram frequency of each noun-property pair in  Google Ngrams \cite{brants2009web}. 
If a noun-property pair does not appear in the corpus, we assign to it a frequency of 0. 

\subsection{Evaluation Metrics}
We evaluate the property ranking proposed by each model using the top-K Accuracy (A@K), top-K recall (R@K), and Mean Reciprocal Rank (MRR) metrics. A@K is defined as the percentage of nouns for which \textbf{at least one} ground truth property is among the top-K predictions  \citep{ettinger-2020-bert}. R@K shows the proportion of ground truth properties retrieved in the top-K predictions. We report the average R@K across all nouns in a test set. MRR stands for the ground truth properties' average reciprocal ranks (more precisely, the inverse of the rank, $\frac{1}{\text{rank}}$). For all three metrics, high scores are better. 

\begin{table*}[!ht]
\centering
\resizebox{16cm}{!}{%
\begin{tabular}{lcc|ccccc|ccccc|ccc}
\Xhline{3\arrayrulewidth}
\multirow{2}{*}{{\bf Model}} & \multirow{2}{*}{{\bf \# Param }} & \multirow{2}{*}{\begin{tabular}[c]{@{}c@{}}
{\bf Img}\end{tabular}}  & \multicolumn{5}{c|}{{\bf {\sc Feature Norms}}}  & \multicolumn{5}{c|}{{\bf {\sc Concept Properties}}-\texttt{test}} & \multicolumn{3}{c}{{\bf {\sc Memory Colors}}}\\ \cline{4-16} 
 & &  & {\bf A@1}     & {\bf A@5}     & {\bf R@5}       & {\bf R@10}      & {\bf MRR}        & {\bf A@1}     & {\bf A@5}     & {\bf R@5}       & {\bf R@10}      & {\bf MRR} & {\bf A@1} & {\bf A@2} & {\bf A@3}       \\ \hline
{\sc Random}     & 0 & \xmark & 1.0 & 2.4 & 0.7 & 1.4 & .018  &  0.2 & 3.8 & 0.5 & 1.7 & .014 & 11.9 & 20.2 & 25.7  \\
{\sc GloVe}   & 0   & \xmark & 16.3 & 42.2 & 16.4 & 26.6 & .124  &  18.5 & 46.6 & 9.5 & 16.4 & .078 & 28.4 & 45.0 & 60.1  \\ 
{\sc Google-Ngram} & 0 & \xmark & 23.4 & 65.2 & 31.5 & 47.7 & .192  &  27.9 & 72.1 & 18.5 & 30.3 & .122 & 44.0 & 63.3 & 69.7 \\\hline
{\sc BERT-large}  & 345M & \xmark & 27.3 & 60.3 & 29.4 & 43.6 & .194  & 31.4 & 72.1 & 18.2 & 29.2 & .123 & 44.0 & 57.8& 67.9 \\
{\sc RoBERTa-large} & 354M & \xmark &  24.6 & 63.1 & 30.2 & 46.3 & .188  & 34.1 & 79.1 & 22.4 & 34.8 & .138 & 48.6 & 61.5 & 67.9 \\
{\sc GPT2-large} & 1.5B & \xmark & 22.0 & 60.7 & 28.4 & 42.9 & .173  &  35.6 & 77.0 & 21.0 & 32.4 & .136 & 44.0 & 57.8 & 67.9 \\
{\sc GPT3-davinci} &  175B & \xmark & 37.9 & 61.5 & 31.8 & 44.2 & -  &  47.0 & 72.2 & 20.1 & 29.7  & -  & 74.3 &82.6 & 84.4\\ \hline
{\sc ViLT}  & 135M & \cmark  &  27.9 & 56.0 & 26.2 & 40.1 & .185  &  34.5 & 63.2 & 15.7 & 23.7 & .118 & 74.3 & -&- \\
{\sc CLIP-ViT/L14} &  427M & \cmark & 28.5 & 61.7 & 29.4 & 42.7 & .197  &  29.2 & 63.0 & 15.0 & 24.9 & .113 & 84.4 & 91.7 & 97.2 \\
 {\sc CEM-Gold} ({\tt GloVe}) & 781M & \cmark & 38.9 & 75.6 & 39.4 & 53.3 & .249  & 48.6 & 84.8 & 27.0 & 39.3 & .171 & 83.5 & 92.7 & \textbf{99.1} \\ 
{\sc CEM-Gold} & 781M & \cmark & \textbf{40.1} & \textbf{76.2} & \textbf{40.0} & \textbf{53.3} & \textbf{.252}  &  48.5 & 84.2 & 26.8 & 38.8 & .170 & 83.5 & 92.7 & \textbf{99.1}\\ 
{\sc CEM-Pred}   & 781M  & \cmark & 39.9 & 75.8 & \textbf{40.0} & 52.5 & .251 & \textbf{49.9} & \textbf{85.8} & \textbf{28.1} & \textbf{40.0} & \textbf{.175 } & \textbf{88.1} & \textbf{96.3} & \textbf{99.1} \\ \Xhline{3\arrayrulewidth}
\end{tabular}
}
\caption{Results obtained on the three 
datasets. The best result for each metric is marked in \textbf{boldface}. \\
}
\label{table: MRD and CSLB main}
\end{table*}

\subsection{Implementation Details}

\label{implementation}
\textbf{Prompt Selection} We evaluate the performance of {\sc BERT-large}, {\sc RoBERTa-large}, {\sc GPT-2-large}, and {\sc ViLT} on the {\tt dev} 
set (cf. Section \ref{datasetssection}) using the prompt templates proposed by   \citet{apidianaki-gari-soler-2021-dolphins}. 
For {\sc CLIP}, we handcraft a set of prompts that are close to the format that was recommended in the original paper \cite{radford2021learning}  and evaluate their performance on the {\tt dev} set. We choose the prompt that yields the highest performance in terms of MRR on the {\tt dev} set for each model, and use it for all our experiments (cf. Appendix \ref{prompt selection} for details). Table \ref{table: final prompt selection} lists the prompt templates selected for each model.
\medbreak
\noindent \textbf{Image Collection} We collect images for the nouns in our datasets using the Bing Image Search API, an image query interface widely used for research purposes~\cite{kiela-etal-2016-comparing, mostafazadeh-etal-2016-generating}.\footnote{We use the \href{https://pypi.org/project/bing-image-downloader/}{bing-image-downloader} API.}
We use again the {\tt dev} set to determine the number of images needed for each noun. We find  that good performance can be achieved with only ten images (cf. Figure \ref{fig:clip size} in Appendix \ref{app: number of images}). Adding more images increases the computations needed without significantly improving the performance. Therefore, we set the number of images per noun to ten for all vision models and experiments.
\medbreak
\noindent \textbf{Model Implementation} All LMs and MLMs are built on the huggingface API.\footnote{\href{https://huggingface.co}{https://huggingface.co}} The {\sc CLIP} model is adapted from the official repository.\footnote{\href{https://github.com/openai/CLIP}{https://github.com/openai/CLIP}} {\sc CEM} ensembles the {\sc RoBERTa-large} and the {\sc CLIP-ViT/L14} models. The experiments were run on Quadro RTX 6000 24GB. All our experiments involve zero-shot and one-shot (for {\sc GPT-3}) probing, hence no training of the models is needed. The inference time of {\sc CEM} is naturally longer than that of individual models, but it is still very fast and only takes a few minutes for each dataset, with pre-computed image features. For more details on runtime refer to Section \ref{app:inference_times},  and specifically to Table \ref{table: inference_times}, in the Appendix.

\begin{table}[!t]
\centering
\resizebox{6.5cm}{!}{%
\begin{tabular}{c|cc}
\Xhline{3\arrayrulewidth}
\multirow{2}{*}{{\bf Noun}}       & \multicolumn{2}{c}{{\bf Property}} \\ \cline{2-3} 
& {\bf most concrete} & {\bf least concrete} \\ \hline
dandelion & yellow & annoying   \\
cougar & brown & vicious  \\
wand & round & magical    \\    
spear & sharp & dangerous \\
pyramid & triangular & mysterious \\ \Xhline{3\arrayrulewidth}
\end{tabular}
}
\caption{Examples of nouns with their most and least concrete properties in {\sc Feature Norms}.}
\label{table: concrete examples}
\end{table}

\section{Evaluation}
\label{sec:evaluation}
\subsection{Property Ranking Task}
\label{sec:task}

Table \ref{table: MRD and CSLB main} shows the results obtained by the LMs, the MLMs and our {\sc CEM} model on the {\sc Feature Norms},  {\sc Concept Properties}-{\tt test}\footnote{Contains all nouns in  {\sc Concept Properties} except from the ones in  the {\sc Concept Properties}-{\tt dev} set.} and {\sc Memory Colors} datasets. The two flavors of {\sc CEM} ({\sc CEM-Pred} and {\sc CEM-Gold}) outperform all other models with a significant margin across datasets.  Interestingly, {\sc CEM-Pred} performs better than {\sc CEM-Gold} on the {\sc Concept Properties}-{\tt test} dataset. This may be due to the fact that 49 properties in this dataset do not have ground truth concreteness scores (vs. only 15 properties in {\sc Feature Norms}), indicating that the prediction model   
probably approximates concreteness better in these cases, contributing to  higher scores for {\sc CEM-Pred}. 

As explained in Section \ref{thecemmodel}, we explore two different heuristics to select the score for these properties for {\sc CEM-Gold}: longest matching subsequence and {\tt GloVE} cosine similarity. The latter similarity metric results to a drop in performance on {\sc Feature Norms} 
and almost identical performance for {\sc Concept-Properties}-{\tt test}.\footnote{Specifically, for A@1 we observe a drop of 1.2 in {\sc Feature Norms} and a gain of .1 for {\sc Concept Properties}}

\begin{figure}[!t]
    \includegraphics[width=7.7cm]{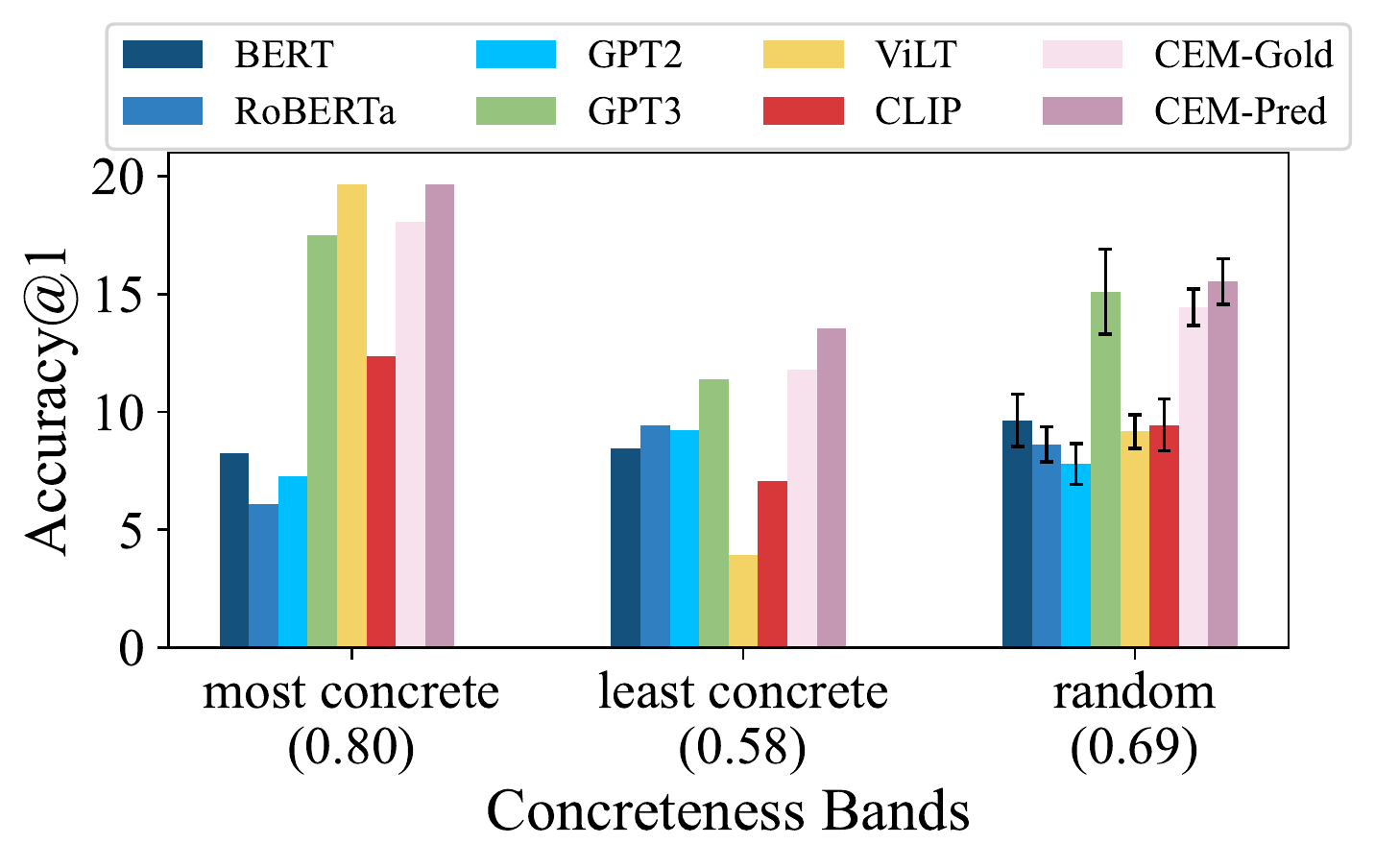}
    \caption{Top-1 Accuracy for the {\sc Feature Norms} properties filtered by concreteness. The average concreteness score for each band is given on the x-axis. The error bars in the ``random'' category represent the standard deviation on 10 trials.} 
    \label{fig:mrd_concrete}
\end{figure}

We notice that the {\sc Google-Ngram } baseline performs well on {\sc Feature Norms} with results on par or superior to big LMs. The somewhat lower results obtained on {\sc Concept Properties}-{\tt test} might be due to the higher number of properties in this dataset (cf. Table \ref{table: dataset stat}), which makes the ranking task more challenging.\footnote{The mean number of properties per noun in  {\sc Concept Properties} is 6.6, 
and  3.1 in {\sc Feature Norms}.}

There is also a higher number of noun-property pairs that are not found in Google Bigrams which are assigned a zero score.\footnote{~26\% of the pairs in {\sc Concept Properties} vs. 15\% for {\sc Feature Norms}.}

The {\sc Memory Colors} dataset associates each noun with a \textit{single} color so we only report Accuracy at top-K (last three columns of Table \ref{table: MRD and CSLB main}). We can compare these scores to a previous baseline, the top-1 Accuracy reported by \citet{norlund-etal-2021-transferring} for the {\sc CLIP-BERT}  model which is 78.5.\footnote{We cannot calculate the other scores because {\sc CLIP-BERT} has not been made available. In this model, a {\sc CLIP} encoded image is appended to {\sc BERT}'s tokenized input before fine-tuning 
with a masked language modeling objective on 4.7M captions paired with 2.9M images. For more details refer to \cite{norlund-etal-2021-transferring}.}

{\sc CEM-Pred} and {\sc Gold} both do better on this dataset (88.1). 
{\sc GPT-3} gets much higher scores than 
the other three language models 
on this task with a top-1 Accuracy of 74.3, but is outperformed by {\sc CLIP} and {\sc CEM}. 
Note that MRR does not apply to {\sc GPT-3} since it generates properties instead of reranking them 
(cf. Appendix \ref{app:gpt3_prompt}).

\begin{figure}[!t]
\centering
    \includegraphics[width=7.3cm]{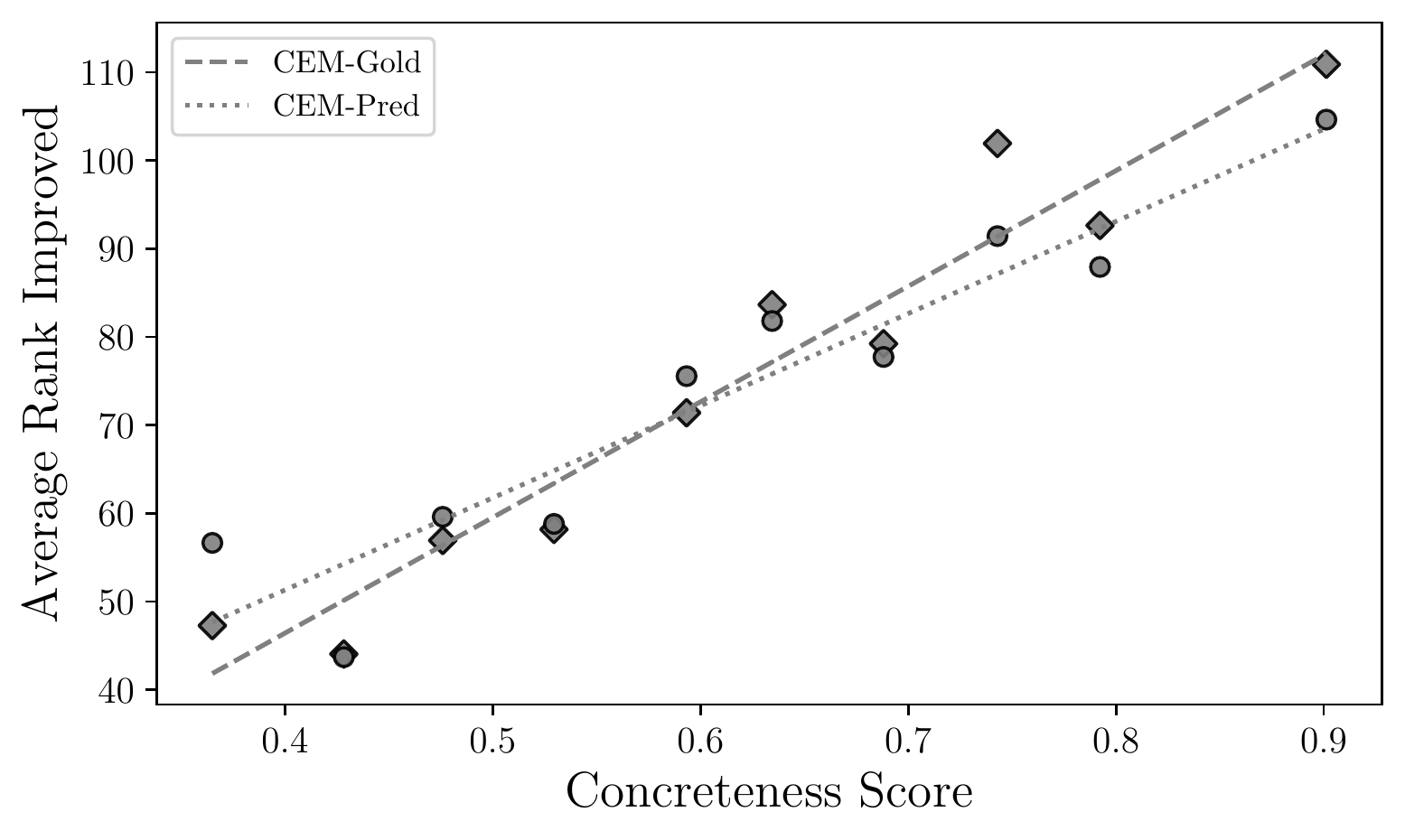}
    \caption{The average Rank Improvement (RI) score for properties in the {\sc Concept Properties}-{\tt test} grouped in ten bins according to their concreteness. The higher the concreteness score of the properties in a bin, the larger the improvement brought by {\sc CEM-Gold} and {\sc CEM-Pred} over {\sc RoBERTa}.}
    \label{fig:cslb_concrete_rank}
\end{figure}

The multimodal model with the lowest performance, {\sc ViLT}, is as good as {\sc GPT-3}. 
{\sc CLIP} falls halfway between {\sc ViLT} and {\sc CEM-Pred/Gold}. 
{\sc CEM-Pred} and {\sc CEM-Gold} present a clear advantage compared to language and multimodal models,  achieving a top-1 Accuracy of 88.1. Although {\sc RoBERTa} gets very low Accuracy on {\sc Memory Colors}, it does not hurt performance when combined with {\sc CLIP} in our {\sc CEM-Gold} model.
This is because the color properties in this dataset have high concreteness scores (0.82 on average), so {\sc CEM-Gold} relies mainly on {\sc CLIP}  which works very well in this setting. {\sc CEM-Gold} makes the same top-1 predictions as {\sc CLIP} for 95 nouns (out of 109), while only 50 nouns are assigned the same color by {\sc CEM-Gold} and  {\sc RoBERTa}. 

\begin{figure}[!t]
\centering
    \includegraphics[width=7.6cm]{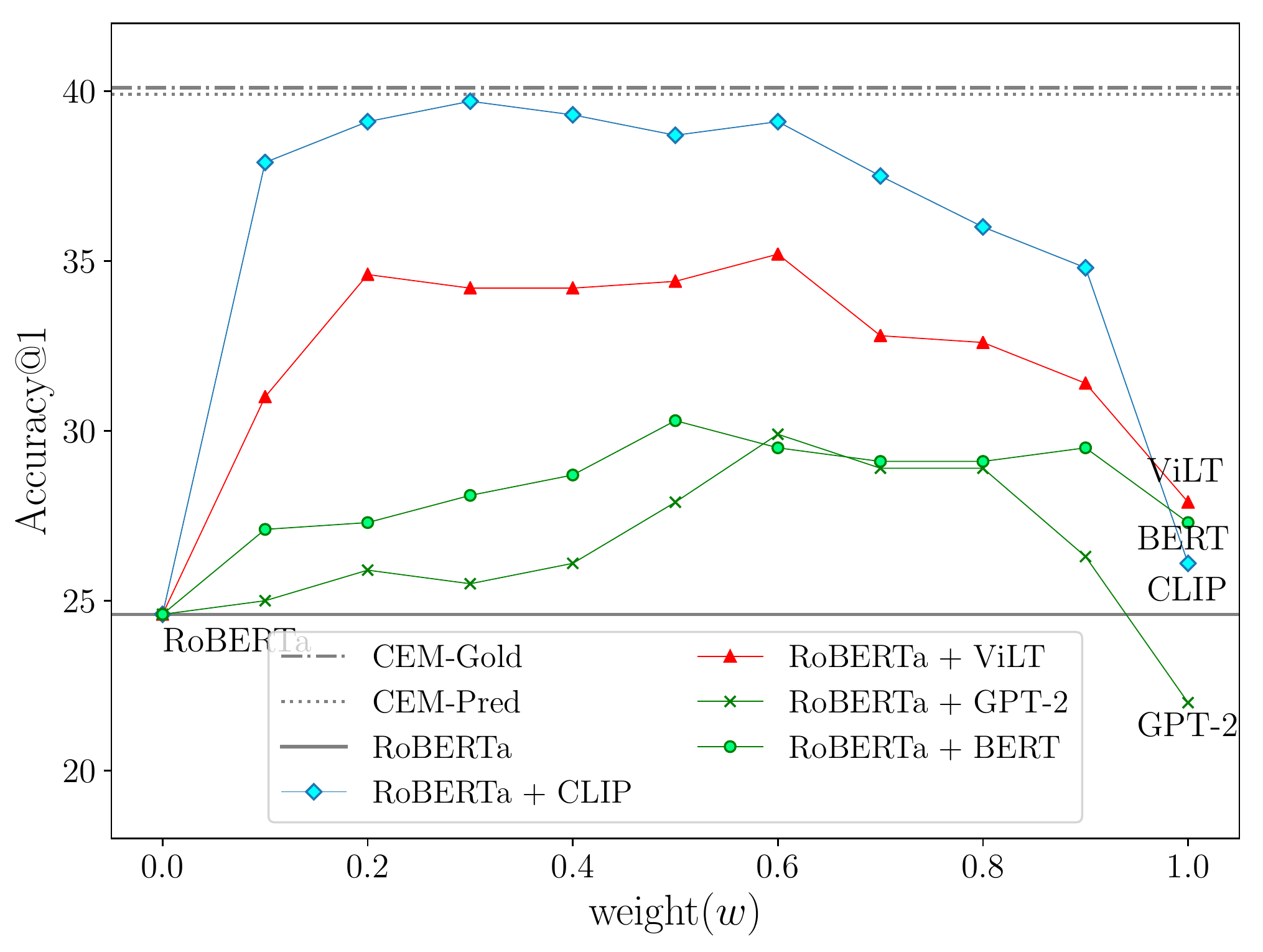}
    \caption{Top-1 Accuracy obtained by different ensemble models on the {\sc Feature Norms}. The x-axis shows the weight used to interpolate two models. The straight dashed and dotted lines are the top-1 Accuracy of {\sc CEM-Gold} (40.1) and {\sc CEM-Pred} (39.9) respectively. }
    \label{fig:combine_model}
\end{figure}

\subsection{Additional Analysis}
\noindent {\bf Concreteness level.} 
We examine the performance of each model for properties at different concreteness levels. From the properties available for a noun in {\sc Feature Norms},\footnote{In this experiment, we use 411 nouns (out of 509) from {\sc Feature Norms} which have at least two properties.} we keep a single property as our ground truth for this  experiment: (a) {\bf most concrete}: the property with the highest concreteness score in the \citet{brysbaert2014concreteness} lexicon; (b) {\bf least concrete}: the property with the lowest concreteness score; 
\noindent(c) {\bf random}: a randomly selected property.\footnote{We report the mean and standard deviation on 10 trials.}
Figure \ref{fig:mrd_concrete} shows the top-1 Accuracy of the models for the properties in each concreteness band. 
Examples of nouns with their most and least concrete properties are given in Table \ref{table: concrete examples}. The results of this experiment confirm our initial assumption that MLMs (e.g., {\sc CLIP} and {\sc ViLT}) are better at capturing concrete properties, and LMs (e.g., {\sc RoBERTa} and  {\sc GPT-2}) are better at identifying abstract ones. {\sc GPT-3} is the only LM that performs better for concrete than for abstract properties, while still falling behind {\sc CEM} variations.

\begin{table}[]

\scalebox{0.8}{
\begin{tabular}{ccc}
\Xhline{3\arrayrulewidth}
\textbf{Noun}     & \textbf{Model}   & \textbf{Top-3 Properties}  \\ \hline

swan & {\sc RoBERTa} &  male, white, black \\
\multirow{4}{*}{\cincludegraphics[width=1.9cm]{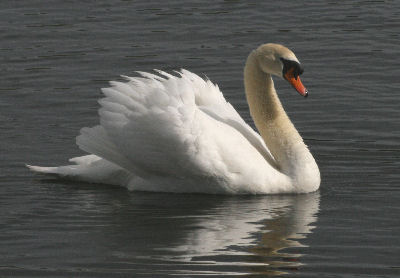}}                          & {\sc CLIP }   &  white, graceful, gentle     \\
                          & {\sc GPT-3 }   &  graceful, regal, stately \\
                        &{\sc CEM-Gold}     & white, large, graceful \\
                        &{\sc CEM-Pred}     & white, endangered, graceful \\\hline

ox & {\sc RoBERTa} &  male, white, black \\
\multirow{4}{*}{\cincludegraphics[width=1.7cm]{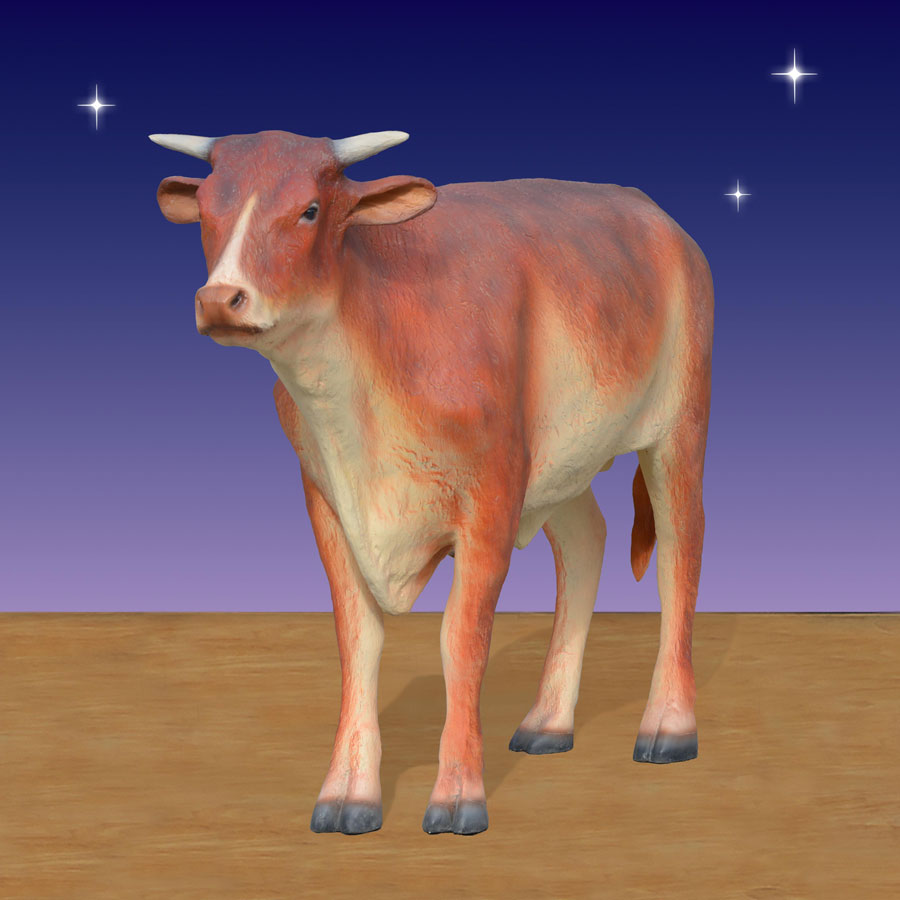}}                          & {\sc CLIP }   &  endangered, wild, harvested     \\
                          & {\sc GPT-3 }   &  strong, muscular, brawny\\
                          &{\sc CEM-Gold}     & large, wild, friendly \\
                           &{\sc CEM-Pred}     & large, wild, hairy \\\hline

plum & {\sc RBTa} & edible, yellow, red  \\
\multirow{3}{*}{\cincludegraphics[width=1.3cm]{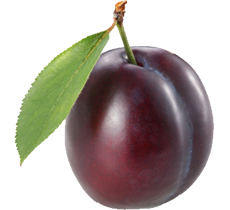}}                           & {\sc CLIP }   &  purple, edible, picked    \\
                          &{\sc CEM}     & edible, purple, harvested \\
                          & {\sc GPT-3 }   & tart, acidic, sweet  \\  \hline

orange & {\sc RoBERTa} &  edible, yellow, orange  \\
\multirow{4}{*}{\cincludegraphics[width=1.9cm]{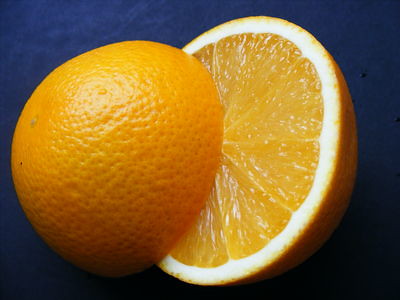}}                           & {\sc CLIP }   &  orange, citrus, juicy  \\
                          & {\sc GPT-3 }   &  tart, acidic, sweet \\ 
                            &{\sc CEM-Gold}     & orange, edible, healthy \\
                          & {\sc CEM-Pred} & orange,edible,citrus \\ \hline

cape & {\sc RoBERTa} & black, white, fashionable \\
\multirow{4}{*}{\cincludegraphics[width=1.9cm]{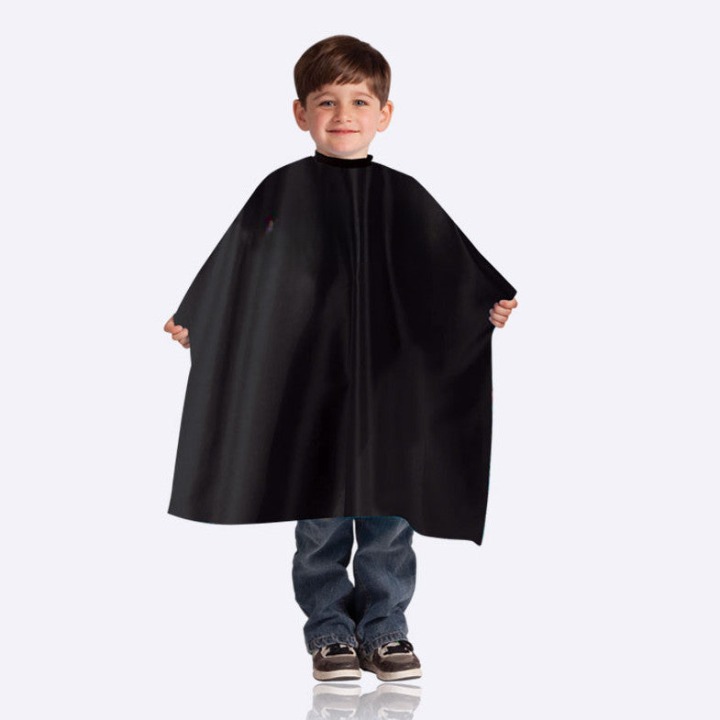}}                           
& {\sc CLIP }   &  cozy, dressy, cold \\
                          & {\sc GPT-3 }   &  tart, acidic, sweet \\
                          & {\sc CEM-Gold}   & fashionable, dark, grey \\
                          &  {\sc CEM-Pred} &  fashionable,grey,dark \\ 
                          
                          \Xhline{3\arrayrulewidth}

\end{tabular}
}
\caption{Top-3 properties proposed by different models for  nouns in {\sc Feature Norms}.}
\label{table:qualitative_results}
\end{table}
\paragraph {\bf Rank Improvement.} We investigate the relationship between the performance of {\sc CEM} and the concreteness score of the properties in {\sc Concept Properties}-{\tt test}. We measure the rank improvement (RI) for a property ($\mathcal{P}$) that occurs when using {\sc CEM} compared to when {\sc RoBERTa} is used as follows: 
\begin{equation} \label{eq: rank improved}
    \text{RI}(\mathcal{P}) = \text{Rank}_{\text{CEM}}(\mathcal{P}) - \text{Rank}_{\text{RoBERTa}}(\mathcal{P})
\end{equation}
\noindent A high RI score for  $\mathcal{P}$ means that its rank is improved with {\sc CEM} compared to {\sc RoBERTa}. We calculate the RI for properties at  different concreteness levels. We  sort the 400 properties in  {\sc Concept Properties}-{\tt test} by increasing the concreteness score and group them into ten bins of 40 properties each. We find a clear positive relationship between the average RI and concreteness scores within each bin, as shown in Figure \ref{fig:cslb_concrete_rank}. This confirms that both {\sc CEM-Pred} and {\sc CEM-Gold} perform better with  concrete properties.

\paragraph {\bf Ensemble Weight Selection.} 
We explore whether a dynamic concreteness-based ensemble weight outperforms a fixed one. We experiment with different model combinations ({\sc RoBERTa} with {\sc BERT}, {\sc GPT-2}, and {\sc ViLT}) with an interpolation weight $w$ that takes values in the range [0,1]. If the weight is close to 0, {\sc CEM} relies more on {\sc RoBERTa}; if it is 1, {\sc CEM} relies more on the second model.
\begin{equation} \label{eq: combine model}
\begin{split}
    \text{Rank}_{\text{combine}}(\mathcal{P}) = (1 &- w) \cdot \text{Rank}_{\text{RoBERTa}}(\mathcal{P}) \\
    &+  w \cdot \text{Rank}_{\text{other model}}(\mathcal{P})
\end{split}
\end{equation}

\noindent We also run the best performing {\sc RoBERTa} + {\sc CLIP} combination again using weights fixed in this way, i.e. without recourse to the properties' concreteness score as in {\sc CEM-Pred} and in {\sc CEM-Gold}. Note that we do not expect the combination of two text-based LMs to improve Accuracy a lot compared to {\sc  RoBERTa} alone. Our intuition is confirmed by the results obtained on {\sc Feature Norms} and shown in Figure \ref{fig:combine_model}. 

The dashed and dotted straight lines in the figure represent the top-1 Accuracy of {\sc CEM-Gold} and {\sc CEM-Pred}, respectively, 
when the weights used are not the ones on the x-axis, but the gold and predicted concreteness scores  (cf. Equation \ref{rankequation}). To further highlight the importance of concreteness in interpolating the models, we provide additional results and comparisons 
in Appendix \ref{app:cem-ablation}. Note that {\sc CEM-Gold} and {\sc CEM-Pred} have highly similar performance and actual output.  

On average over all nouns, they propose 4.35 identical properties at top-5 for nouns in {\sc Feature Norms}, and 4.41 for nouns in {\sc Concept Properties}-{\tt test}. 

\begin{figure}[!t]
\centering
    \includegraphics[width=\linewidth]{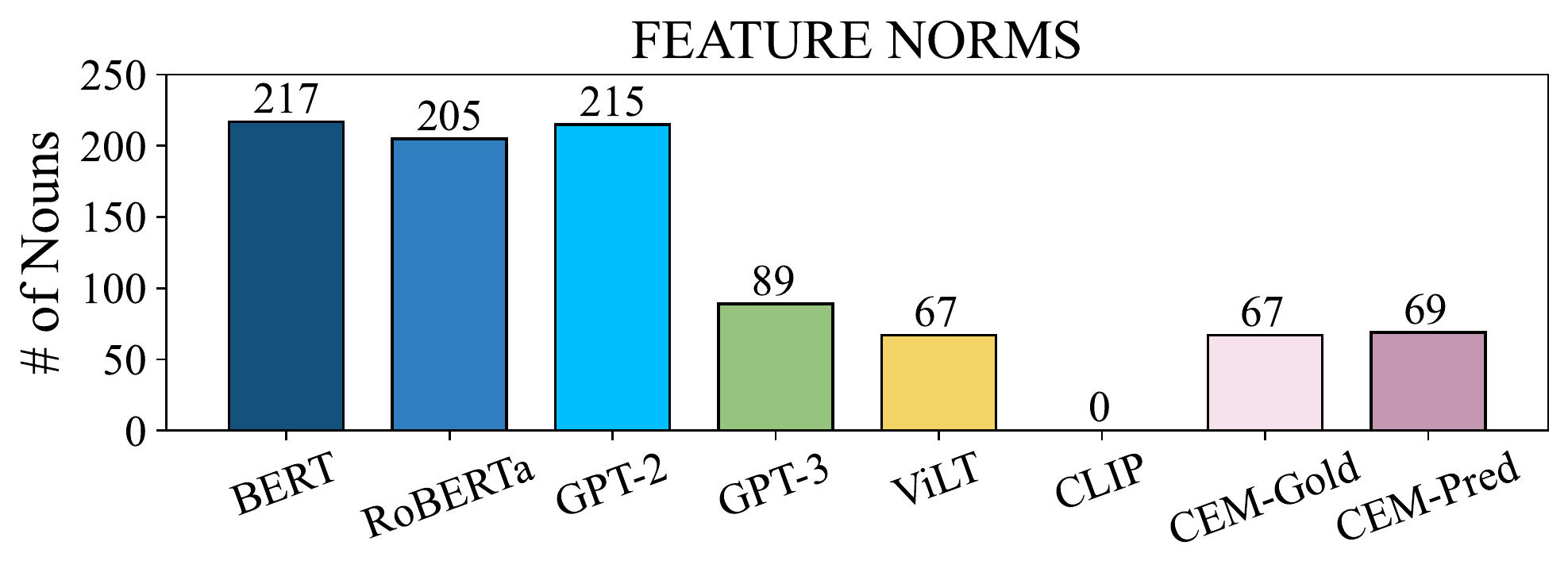}
    \includegraphics[width=\linewidth]{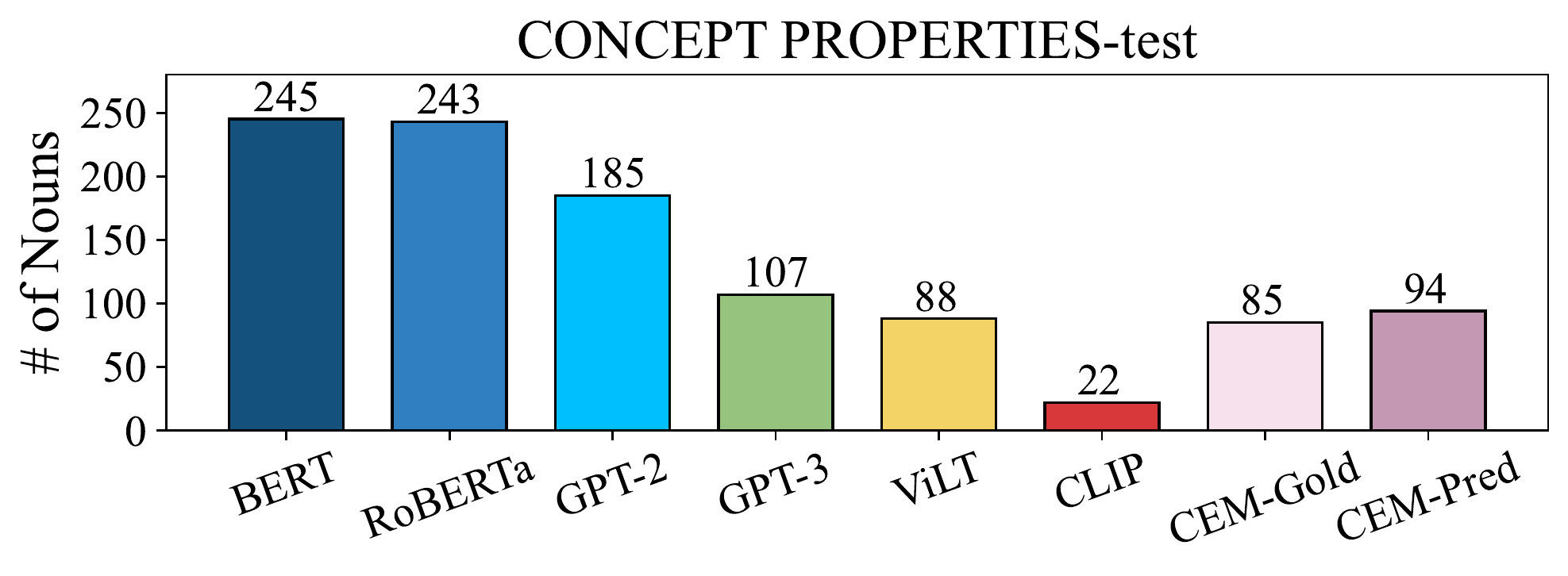}
    \caption{Number of nouns in  {\sc Feature Norms} and {\sc Concept Properties}-{\tt test}  for which a model proposed the same top-3 properties in the same order.}
    \label{fig:duplicate_predictions}
\end{figure}

We observe a slight improvement in top-1 Accuracy (5\%) when ensembling two text-based LMs ({\sc RoBERTa} + {\sc BERT} or {\sc RoBERTa} + {\sc GPT-2}). Text-based LMs have similar output distributions, hence combining them does not change the final distribution much. 
The {\sc RoBERTa} + {\sc ViLT} ensemble model achieves higher performance due to the interpolation with an image-based model, but it does not  reach the Accuracy of the {\sc CEM} models ({\sc RoBERTa} + {\sc CLIP}). The {\sc ViLT} model gets lower performance than {\sc CLIP} when combined with {\sc RoBERTa},  because it was exposed to much less data than {\sc CLIP} during training (400M vs.~30M).
Finally, we notice that the best performance of {\sc RoBERTa} + {\sc CLIP} with fixed weight is slightly lower than that of the {\sc CEM} models. This indicates that using a fixed weight to ensemble two models hurts performance compared to calibrating their mutual contribution using the concreteness score. 
Another advantage of the concreteness score is that it is more transferable since it does not require tuning on new datasets. 
\paragraph{Properties Quality.} Table \ref{table:qualitative_results} shows a random sample of the top-3 predictions made 
by each model for nouns in  {\sc Concept Properties}-{\tt test}. We notice that the properties proposed by the two flavors of {\sc CEM} 
are both perceptual and abstract, due to their access to both a language and a vision model. 
We further observe that {\sc CEM} retrieves rarer and more varied properties for different nouns, compared to the language models.
\footnote{Details on the frequency of the properties 
retrieved by each model are reported in Appendix \ref{app: prediction_frequency}. We provide more randomly sampled qualitative examples in Appendix \ref{app: more_qual}.}

Figure \ref{fig:duplicate_predictions} shows the number of nouns for which a model made the exact same top-3  predictions.\footnote{Refer to \ref{app:duplicates} for the number of nouns with exact same top-K predictions for different values of K.} For example {\sc GPT-3} proposed the properties [{\it tart, acidic, sweet, juicy, smooth}] for 20 different nouns\footnote{apple, plum, grapefruit, tangerine, orange, lime, lemon, grape, rhubarb, cherry, cap, cape, blueberry, strawberry, pine, pineapple, prune, raspberry, nectarine, cranberry} in the same order. Note that better prompt engineering might decrease the number of repeated properties. However, we are already prompting {\sc GPT-3} with one shot, whereas the other models, including {\sc CEM} are zero-shot.
{\sc RoBERTa} predicted [{\it male, healthy, white, black, small}] for  both  {\it mittens} and {\it penguin}, and [{\it male, black, white, brown, healthy}] for {\it owl} and {\it flamingo}. We observe that {\sc CEM-Pred} and {\sc CEM-Gold} 
are less likely to retrieve the same top-K predictions for a noun   than  language models. 
{\sc CEM} combines the variability and accuracy of {\sc CLIP} with the benefits of text-based models, which are exposed to large volumes of texts during pre-training.

\vspace{-0.08cm}
\section{Conclusion}
\vspace{-0.1cm}
We propose a new ensemble model for noun property prediction which leverages the strengths of language models and multimodal (vision) models.

Our model, CEM, calibrates the contribution of the two types of models in a property ranking task by relying on the properties' concreteness level. 
The results show that the CEM model, which combines  
{\sc RoBERTa} and {\sc CLIP} outperforms  powerful text-based language models (such as {\sc GPT-3}) 
with significant margins in three evaluation datasets. 
Additionally, our methodology yields better performance than  alternative ensembling techniques, confirming our hypothesis that 
 concrete properties are more accessible through images, and abstract properties 
through text. 
The Accuracy scores obtained on the larger datasets 

show that there is still room for improvement for this challenging task. 

\section{Limitations}
Our experiments address concreteness at the lexical level, specifically using scores assigned to adjectives in an external resource \cite{brysbaert2014concreteness} or predicted using \cite{charbonnier2019predicting}. 
Another option would be to use the concreteness of the noun phrases formed by the adjectives and the nouns they modify.
We would expect this to be different than the concreteness of adjectives in isolation since the concreteness of the nouns would have an impact on that of the resulting phrase  

(e.g., {\it useful knife} vs. {\it useful idea}). 
We were not able to evaluate the impact of noun phrase concreteness on property prediction 
because the property datasets  used in our experiments mostly contain concrete nouns. 
Another limitation of our methodology is the reliance on pairing images with nouns. In particular, we use a  search engine to retrieve images corresponding to nouns in order to get grounded predictions from the vision model. 
Finally, we only evaluate our methodology in English and leave experimenting with other  languages to future work, since this would require the collection of multi-lingual semantic association datasets and/or the translation of existing ones. We did not pursue this extension for this paper as \href{https://github.com/FreddeFrallan/Multilingual-CLIP}{\sc Multilingual Clip} model weights only became available very recently. 

\section{Acknowledgements}
We thank Marco Baroni for his feedback on an earlier version of the paper. 
This research is based upon work supported in part by the DARPA KAIROS Program (contract FA8750-19-2-1004), the DARPA LwLL Program (contract FA8750-19-2-0201), the IARPA BETTER Program (contract 2019-19051600004), and the NSF (Award 1928631). Approved for Public Release, Distribution Unlimited. The views and conclusions contained herein are those of the authors and should not be interpreted as necessarily representing the official policies, either expressed or implied, of DARPA, IARPA, NSF, or the U.S. Government.

\bibliography{anthology,custom}
\bibliographystyle{acl_natbib}

\clearpage
\appendix

\section{Prompt Selection} \label{prompt selection}
\subsection{Language Model Prompts} \label{lm prompts}

In our experiments with language models, we use the 11 
prompts proposed  by~\citet{apidianaki-gari-soler-2021-dolphins} for retrieving noun properties. As shown in Table \ref{table: lm prompts}, these involve nouns in singular and plural forms. The performance achieved by each language model with these prompts on the  {\sc Concept Properties} development set  is given in Table \ref{table:lm prompt selection}. The results show that  model  performance varies significantly with different prompts. The best-performing prompt is different for each model. For {\sc BERT} and {\sc GPT-2}, the ``most + PLURAL'' obtains the highest Recall and MRR scores. The best performing prompt for {\sc RoBERTa-large} is ``SINGULAR + generally'', and ``PLURAL'' for {\sc ViLT}.
\begin{table}[h]
\centering
\resizebox{7.5cm}{!}{%
\begin{tabular}{ll}
\Xhline{3\arrayrulewidth}
\multicolumn{1}{c}{Prompt Type} & \multicolumn{1}{c}{Prompt Example} \\ \hline
SINGULAR    &   a motorcycle is \texttt{[MASK]}.              \\
PLURAL      &   motorcycles are \texttt{[MASK]}.              \\
SINGULAR + usually  &  a motorcycle is usually \texttt{[MASK]}.  \\
PLURAL + usually    &  motorcycles are usually \texttt{[MASK]}.               \\
SINGULAR + generally    &  a motorcycle is generally \texttt{[MASK]}.               \\
PLURAL + generally  &  motorcycles are generally \texttt{[MASK]}.               \\
SINGULAR + can be   &  a motorcycle can be \texttt{[MASK]}.               \\
PLURAL + can be &   motorcycles can be \texttt{[MASK]}.              \\
most + PLURAL   &   most motorcycles are \texttt{[MASK]}.              \\
all + PLURAL    &  all motorcycles are \texttt{[MASK]}.               \\
some + PLURAL   & some motorcycles are \texttt{[MASK]}.                \\ \Xhline{3\arrayrulewidth}
\end{tabular}
}
\caption{Prompts used for language models.}
\label{table: lm prompts}
\end{table}

\subsection{{\sc CLIP} Prompts}
For {\sc CLIP}, we handcraft ten prompts and report their performance on the  {\sc Concept Properties} development set  in Table \ref{table: clip prompts}. Similar to what we observed with language models, {\sc CLIP} performance is also sensitive to the prompts used. We select for our experiments the prompt ``An object with the property of \texttt{[MASK]}.'', which obtains the highest average Accuracy and MRR score 
on the {\sc Concept Properties}  development set.
\begin{table}[h]
\centering
\resizebox{7.5cm}{!}{%
\begin{tabular}{ccccc}
\Xhline{3\arrayrulewidth}
Prompt Type            & Acc@1 & R@5  & R@10 & MRR\\ \hline
\texttt{[MASK]}             &  26.0 & 13.1 & \textbf{21.9} & .097   \\
This is \texttt{[MASK]}.    &  28.0 & 9.6 & 13.6 & .089   \\
A \texttt{[MASK]} object.   &  22.0 & 13.2 & 18.9 & .089  \\
This is a \texttt{[MASK]} object. & 22.0 & 12.0 & 17.2 & .087    \\
The item is \texttt{[MASK]}. &  18.0 & 7.5 & 17.2 & .074  \\
The object is \texttt{[MASK]}.                & 24.0 & 10.5 & 16.2 & .088    \\
The main object is \texttt{[MASK]}.           & 24.0 & 10.3 & 20.3 & .091    \\
An object which is \texttt{[MASK]}.           & 28.0 & \textbf{13.7} & 19.9 & .106    \\
An object with the property of \texttt{[MASK]}.           & \cellcolor{gray!30}\textbf{32.0} & \cellcolor{gray!30}12.3 & \cellcolor{gray!30}20.0 & \cellcolor{gray!30}\textbf{.108}    \\\Xhline{3\arrayrulewidth}
\end{tabular}
}
\caption{Full results of CLIP-ViT/L14 on the {\sc Concept Properties} development set.}
\label{table: clip prompts}
\end{table}
\begin{table*}[!t]
\centering
\resizebox{16cm}{!}{%
\begin{tabular}{c|ccc|ccc|ccc|ccc}
\Xhline{3\arrayrulewidth}
\multirow{2}{*}{Prompt Type} & \multicolumn{3}{c|}{BERT-large} & \multicolumn{3}{c|}{RoBERTa-large} & \multicolumn{3}{c|}{GPT-2-large} & \multicolumn{3}{c}{ViLT} \\ \cline{2-13} 
& R@5 & R@10 & MRR     & R@5 & R@10 & MRR & R@5 & R@10 & MRR & R@5    & R@10    & MRR   \\ \hline
SINGULAR               &   8.9   &  17.3  &  .067  &     17.1  &  23.6  &  .092    &    14.0  &  27.5  &  .097    &  12.6  &  18.2  &  .085 \\
PLURAL                &   11.5  &  21.9  &  .070  &     10.5  &  21.1  &  .085    &    14.9  &  23.7  &  .098    &  \cellcolor{gray!30}15.5  &  \cellcolor{gray!30}\textbf{24.5}  &  \cellcolor{gray!30}\textbf{.105}\\
SINGULAR + usually     &   12.7  &  24.5  &  .082  &    15.5 & 26.5 & .098     &    16.2  &  25.3  &  .107    &  11.8  &  18.7  &  .088 \\
PLURAL + usually       &   14.4  &  \textbf{27.6}  &  \textbf{.107}  &     13.3  &  23.7  &  .106    &    17.8  &  24.6  &  .113    &  \textbf{15.6}  &  21.7  &  .091 \\
SINGULAR + generally   &   14.3  &  23.6  &  .087  &    \cellcolor{gray!30}\textbf{17.7}  &  \cellcolor{gray!30}\textbf{27.9}  &  \cellcolor{gray!30}\textbf{.119}     &    18.7  &  29.2  &  .114    &  12.7  &  19.4  &  .083 \\
PLURAL + generally     &   15.0  &  26.7  &  .097  &     16.0  &  25.3  &  .105    &    17.4  &  26.7  &  .128    &   9.8  &  18.6  &  .075 \\
SINGULAR + can be      &   12.4  &  23.9  &  .102  &     14.7  &  22.7  &  .090    &    14.3  &  24.7  &  .105    &   9.2  &  14.1  &  .056 \\
PLURAL + can be        &   16.0  &  26.4  &  \textbf{.107}  &     12.1  &  17.7  &  .073    &    10.2  &  18.3  &  .096    &  10.0  &  14.2  &  .060 \\
most + PLURAL          &   \cellcolor{gray!30}\textbf{16.7}  &  \cellcolor{gray!30}27.3  &  \cellcolor{gray!30}\textbf{.107}  &     12.6  &  25.7  &  .098    &    \cellcolor{gray!30}\textbf{20.0}  &  \cellcolor{gray!30}\textbf{33.4}  &  \cellcolor{gray!30}\textbf{.122}    &  12.6  &  20.8  &  .095 \\
all + PLURAL           &   13.4  &  20.5  &  .083  &      8.2  &  13.5  &  .073    &    19.6  &  31.3  &  .113    &  14.4  &  20.4  &  .103 \\
some + PLURAL          &   11.2  &  21.5  &  .082  &     16.4  &  23.5  &  .100    &    15.4  &  31.5  &  .097    &  10.7  &  17.2  &  .091 \\ \Xhline{3\arrayrulewidth}
\end{tabular}
}
\caption{Full results of language models on the {\sc Concept Properties} development set with different prompts. The best scores for each metric are \textbf{bold}. The best prompt for each model is \colorbox{gray!30}{highlighted}, selected based on the average performance over all metrics. }
\label{table:lm prompt selection}
\end{table*}
\begin{table*}[!t]
\resizebox{16cm}{!}{%
\begin{tabular}{c|cccc|cccc|ccc}
\Xhline{3\arrayrulewidth}
           & \multicolumn{4}{c|}{\sc Feature Norms}   & \multicolumn{4}{c|}{{\sc Concept Properties}-{\tt test}}  & \multicolumn{3}{c}{\sc Memory Colors} \\ \cline{2-12} 
                    & Acc@1 & R@5  & R@10 & MRR  & Acc@1 & R@5  & R@10 & MRR  & Acc@1     & Acc@3     & Acc@5     \\ \hline
CLIP-ViT/B32        & 24.8 & 24.8 & 36.1 & .172 & 27.6 & 13.0 & 19.6 & .097 & 83.5 & 95.4 & \textbf{99.1}     \\
CLIP-ViT/B16        & 25.3 & 27.4 & 38.9 & .184 & 28.3 & 14.3 & 22.0 & .103 & \textbf{87.2} & \textbf{96.3} & 98.2      \\ 
CLIP-ViT/L14        & \textbf{26.1} & \textbf{29.2} & \textbf{43.3} & \textbf{.192} & \textbf{29.2} & \textbf{15.0} & \textbf{24.9} & \textbf{.113} & 82.6 & \textbf{96.3} & \textbf{99.1}      \\ \Xhline{3\arrayrulewidth}
\end{tabular}
}
\caption{Performance of CLIP models with different sizes.}
\label{table: clip size}
\end{table*}

\subsection{{\sc GPT-3} Prompts}
\label{app:gpt3_prompt}
Since we do not have complete control of {\sc GPT-3} at this moment, we treat  {\sc GPT-3} as a question-answering model using the following prompt in a one-shot example setting: 
\begin{verbatim}
Use ten adjectives to describe 
the properties of kiwi:\n
1. tart\n2. acidic\n3. sweet\n
4. juicy\n5. smooth\n6. fuzzy\n
7. green\n8. brown\n9. small\n
10. round\n
Use ten adjectives to describe
the properties of [NOUN]:\n
\end{verbatim}
We use the \texttt{text-davinci-001} engine of {\sc GPT-3} which costs \$0.06 per 1,000 tokens. On average, it costs \$0.007 to generate 10 properties for each noun.

\section{Inference Times}
\label{app:inference_times}

Table \ref{table: inference_times} provides details about the runtime of the experiments. The second column of the Table indicates whether a model uses images. Training the concreteness predictor for {\sc CEM-Pred} takes 10 minutes.  Inference for all nouns in the datasets  with {\sc CEM-Pred} only takes a couple of seconds. Note that {\sc CEM-Pred} is faster than {\sc CEM-Gold}, since {\sc CEM-Gold} leverages the longest matching subsequence heuristic ({\tt LMS}) or {\tt GloVe} vector cosine similarity in order to find the concreteness score of the most similar word in \citet{brysbaert2014concreteness} for properties without a gold concreteness score. The times reported in the table for image feature pre-computation correspond to the time needed for computing embeddings for 200 images for each  noun in a dataset, which is only computed once for each dataset. We, however, only use 10 of them for the final {\sc CEM} models (cf. Appendix \ref{app: number of images}). 

\begin{table*}[!ht]
\centering
\resizebox{16cm}{!}{%
\begin{tabular}{ll|ll|ll|ll|}
 &  & \multicolumn{2}{l|}{{\sc Feature Norms}}  & \multicolumn{2}{l|}{{\sc Concept Properties}-{\tt test}} & \multicolumn{2}{l|}{{\sc Memory Colors}}  \\ \cline{3-8} 
Model & Img    & Time & \begin{tabular}{cc}Image Features\\ Pre-Computation \end{tabular}  & Time &  \begin{tabular}{cc}Image Features\\ Pre-Computation \end{tabular}    & Time & \begin{tabular}{cc}Image Features\\ Pre-Computation \end{tabular} \\ \hline
{\sc GloVe}                 & \xmark &  11 sec.    &            \qquad\qquad   -                        &   12 sec.        &    \qquad\qquad   -                                    &     10 sec. &       \qquad\qquad   -                             \\
{\sc Google Ngram}          & \xmark &  15 min.    &                 \qquad\qquad    -                     &   15 min.        &            \qquad\qquad    -                              &    15 min.  &             \qquad\qquad   -                         \\ \hline
{\sc BERT-large}          & \xmark &   3 min. 18 sec.     &   \qquad\qquad   -                    &     7 min 33 sec.      &       \qquad\qquad    -                          &  4 sec.    &  \qquad\qquad    -                           \\
{\sc RoBERTa-large}          & \xmark &  2 min. 31 sec.     &       \qquad\qquad    -                     &     5 min 50 sec      &     \qquad\qquad     -                          &   3 sec.   &  \qquad\qquad    -                           \\
{\sc GPT2-large}             & \xmark &   48 min. 2 sec.  &     \qquad\qquad      -                     &     1 hr. 39 min.       &     \qquad\qquad      -                          &   38 sec.   &   \qquad\qquad -                           \\
{\sc GPT3-davinci}           & \xmark &   6 min. 50 sec.   &    \qquad\qquad       -                     &    8 min 7 sec       &    \qquad\qquad      -                          & 1 min. 27 sec.    &  \qquad\qquad   -                           \\\hline
{\sc ViLT}                   & \cmark &    1 hr. 40 min.  &   \qquad  2 hr. 50 min.                         &  2 hr. 45 min.         &   \qquad  3 hr. 20 min.                             &  57 sec.    &    \qquad\qquad    33 min.                     \\
{\sc CLIP-ViLT/L14}          & \cmark &    52 seconds  &     \qquad         5 hr. 40 min.                 &      2 min. 10 sec.     &       \qquad     6 hr.         41 min.                 &  13 sec.       &              \qquad 1 hr. 13 min                     \\
{\sc CEM-gold}  ({\tt GloVE})            & \cmark &  4 min. 14 sec.    &          \qquad         5 hr. 40 min.              &        10 min. 4 sec.       &         \qquad             6 hr.         41 min.                            &    28 sec.  &              \qquad 1 hr. 13 min                     \\
{\sc CEM-gold}  ({\tt LMS})            & \cmark &  3 min. 30 sec.    &          \qquad         5 hr. 40 min.              &        8 min. 12 sec.      &         \qquad             6 hr.         41 min.                            &      20 sec.  &              \qquad 1 hr. 13 min                     \\
{\sc CEM-pred}               & \cmark &    4 min. 29 sec.  &              \qquad         5 hr. 40 min.                 &     7 min. 20 sec.            &           \qquad       6 hr.         41 min.                               &    49 sec. &          \qquad 1 hr. 13 min                   
\end{tabular}
}
\caption{Experiment inference times. Note that all models are used in zero-shot scenarios with no fine-tuning involved.}
\label{table: inference_times}
\end{table*}


\section{Implementation of {\sc CLIP}}

\subsection{Number of Images}
\label{app: number of images}
\begin{figure}[h]
\centering
    \includegraphics[width=7.5cm]{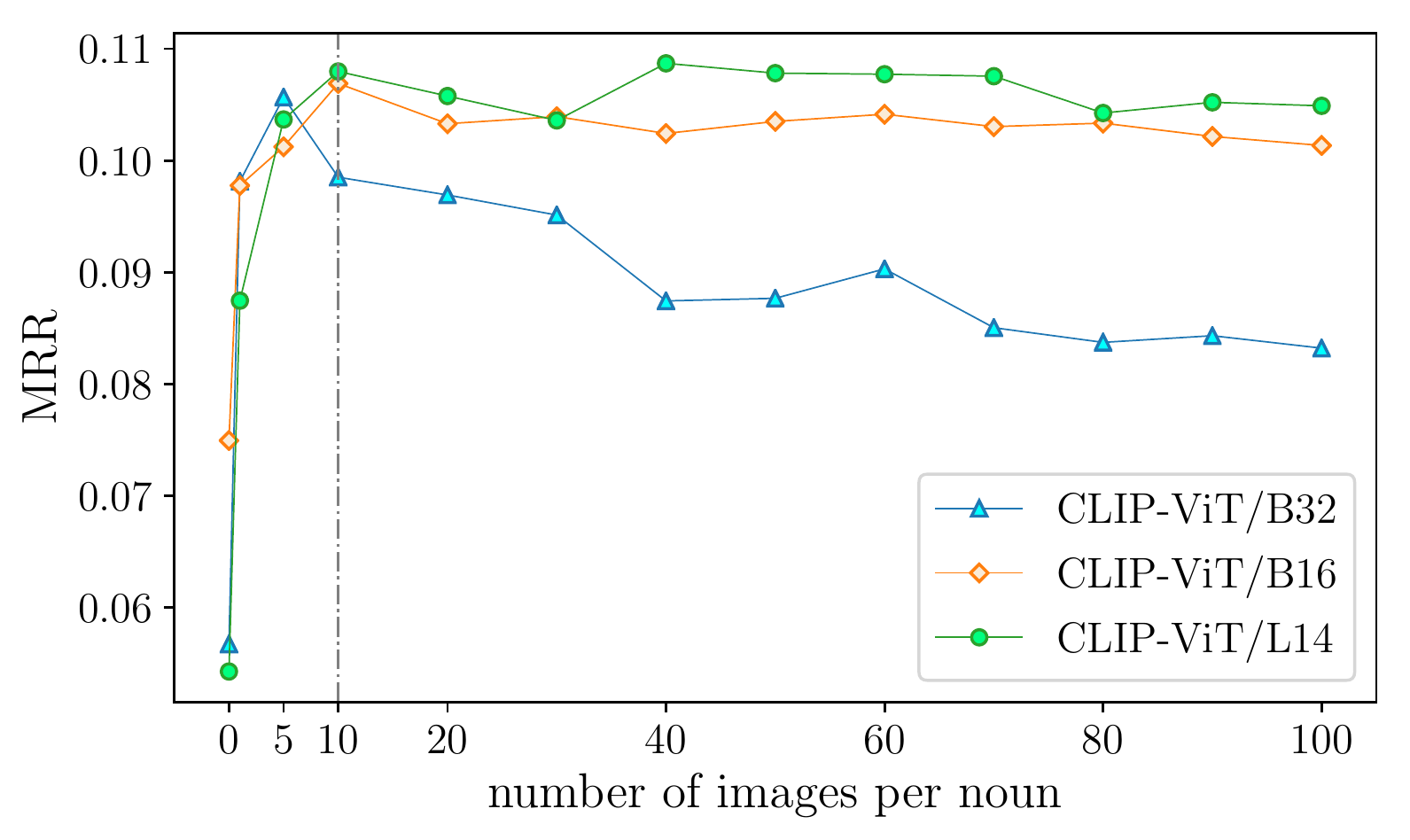}
    \caption{{\sc CLIP} performance on  {\sc Concept Properties}-{\tt test} development set with a different number of images per noun.}
    \label{fig:clip size}
\end{figure}
For each noun, we collected 200 images from 
Bing. Given that it is not practical to use such a high  number of images for a large-scale experiment, we  investigate the performance of {\sc CLIP} with 
different number of images. We first filter the 200 images collected for each noun to remove duplicates. We then sort the remaining images based on the cosine similarity of each image with the sentence ``A photo of \texttt{[NOUN]}.''. 

We pick the top-M images and gradually increase the value of M.\footnote{When M = 0, we use the {\sc CLIP} text encoder to encode the noun as the image embedding.} Figure  \ref{fig:clip size} shows the MRR obtained by {\sc CLIP} on the {\sc Concept Properties} development set with a varying number of images. 
We observe that the model's MRR score increases with a higher number of images. 
Nevertheless, the improvement is marginal when the number of images is higher than ten and starts to overfit when the number is higher 
than 20. Therefore, we decided to use ten images for  all experiments involving {\sc CLIP}.

\subsection{{\sc CLIP} Size}
We evaluate three sizes of {\sc CLIP}, from small to large: {\sc CLIP-ViT/B16}, {\sc CLIP-ViT/B32}, and {\sc CLIP-ViT/L14}. 
As shown in Figure \ref{fig:clip size}, the performance positively correlates with the model size. The largest model, {\sc CLIP-ViT/L14} has a higher MRR score than the other two models. We also report the performance of the three {\sc CLIP} models on {\sc Feature Norms},  {\sc Concept Properties}-{\tt test}, and {\sc Memory Colors} in Table \ref{table: clip size}, indicating that the larger {\sc CLIP} model yields better performance across  metrics.

\begin{table*}[!t]
\centering
\resizebox{16cm}{!}{%
\begin{tabular}{cc|ccccc|ccccc}
\Xhline{3\arrayrulewidth}
\multirow{2}{*}{{\bf Model}} & \multirow{2}{*}{\begin{tabular}[c]{@{}c@{}}
{\bf Images}\end{tabular}}  & \multicolumn{5}{c|}{{\tt{Non-Prototypical}}}  & \multicolumn{5}{c}{{\tt Prototypical}} \\ \cline{3-12} 
 & & {\bf Acc@5}     & {\bf Acc@10}     & {\bf R@5}       & {\bf R@10}      & {\bf MRR}        & {\bf Acc@5}     & {\bf Acc@10}     & {\bf R@5}       & {\bf R@10}      & {\bf MRR}       \\ \hline
{\sc Random}      & \xmark & 4.13 & 7.67 & 2.73 & 4.96 & 0.030 & 4.66 & 8.03 & 2.15 & 3.84 & 0.025   \\
{\sc GloVe}       & \xmark & 22.59 & 33.20 & 16.99 & 26.76& 0.124 & 30.05 & 44.56 & 15.68 & 26.71 & 0.124    \\ 
{\sc Google-Ngram} & \xmark & {\bf 45.19} & {\bf 57.96} & {\bf 39.22} & {\bf 58.80} & {\bf 0.240}  & 39.64  &  56.99 & 24.06 & 36.47 & 0.142    \\\hline
{\sc BERT-large}  & \xmark & 35.76 & 51.28 & 30.22 & 48.12 & 0.197& 45.60 & 58.81 & 28.16 & 39.42 & 0.191   \\
{\sc RoBERTa-large} & \xmark & 35.76 & 48.92 & 28.53 & 46.39 & 0.176 & 47.67 & 63.73 & 28.95  & 43.08 & 0.200    \\
{\sc GPT2-large}  & \xmark & 36.35 & 48.92 & 29.92 & 45.79 & 0.181 & 40.93 & 55.96 & 24.12 & 37.23 & 0.166    \\
{\sc GPT3-davinci}& \xmark & 30.84 & 40.67 & 25.77 & 39.42 & - & 55.18 & 64.51 & 38.30 & 49.66 & -  \\ \hline
{\sc ViLT}  & \cmark & 34.97 & 46.76 & 28.85 & 42.70 & 0.211 & 38.34 & 53.63 & 23.52 & 36.57 & 0.159   \\
{\sc CLIP-ViT/L14} & \cmark & 32.22 & 43.81 & 25.08 & 37.95 & 0.159 & 52.59 & 69.95 & 33.67 & 49.82 & 0.226   \\
{\sc CEM-Gold} (Ours)  & \cmark & 41.85 & 54.03 & 35.88 & 49.55 & 0.217 & { 64.77} & {\bf 75.39} & { 43.11} & { 56.06} & {0.289 } \\
{\sc CEM-Pred} (Ours)  & \cmark & 41.65 & 51.47 & 35.11 & 46.46 & 0.211 & {\bf 65.80} & {74.87} & {\bf 44.67} & {\bf 56.20} & {\bf 0.306 } \\
\Xhline{3\arrayrulewidth}
\end{tabular}
}
\caption{Results obtained on the {\sc Feature Norms} dataset filtered by prototypical and non-prototypical properties. The splits are derived from \cite{apidianaki-gari-soler-2021-dolphins}. }
\label{table: MRD prototypical}
\end{table*}

\begin{table*}[!t]
\resizebox{16cm}{!}{%
\begin{tabular}{c|cccc|cccc|ccc}
\Xhline{3\arrayrulewidth}
           & \multicolumn{4}{c|}{\textbf{{\sc Feature Norms}}}   & \multicolumn{4}{c|}{\textbf{{\sc Concept Properties}{\tt-test}}}  & \multicolumn{3}{c}{\textbf{{\sc Memory Colors}}} \\ \cline{2-12} 
           & \textbf{Acc@1} & \textbf{R@5}  & \textbf{R@10} & \textbf{MRR}  & \textbf{Acc@1} & \textbf{R@5}  & \textbf{R@10} & \textbf{MRR}  & \textbf{Acc@1}     & \textbf{Acc@3}     & \textbf{Acc@5}     \\ \hline
{\sc CEM-Gold }      & \textbf{40.1}  & \underline{40.5} & \textbf{53.3} & \textbf{.252} & 48.3  & 26.9 & 39.1 & .171 & \underline{82.6} & \textbf{96.3} & \textbf{99.1}     \\
{\sc CEM-Pred }  & \underline{39.9}  & 40.4 & 52.5 & .\underline{251} & \textbf{49.9}  & \textbf{28.1} & \underline{40.0} & \textbf{.175} & \textbf{84.4} & 97.2 & \textbf{99.1}     \\
{\sc CEM-random} & 35.4 & 38.3 & 51.0 & .232  & 46.3  & 25.3 & 36.5 & .162 & 62.4 & 90.8 & 94.5      \\ 
{\sc CEM-average} & 38.7 & \textbf{41.0} & \underline{53.0} & .249 & 48.3  & \underline{28.0} &\textbf{40.2} & \underline{.173} & 71.6 & 92.7 & \textbf{99.1}      \\
{\sc CEM-max }    & 36.9 & 38.4 & 51.3 & .238 & \underline{48.6}  & 26.7 & 38.1 & .167 & 67.0 & 90.8 & 96.3      \\
{\sc CEM-min}     & 25.1 & 34.2 & 50.1 & .204 & 30.1  & 21.2 & 34.1 & .135 & 69.7 & \underline{95.4} & \underline{98.2}  \\\Xhline{3\arrayrulewidth}
\end{tabular}
}
\caption{Comparison of ensemble methods on the three datasets. The highest score for each metric is \textbf{bolded} and the second-best is \underline{underlined}.}
\label{table: cem ablation}
\end{table*}

\section{{\sc CEM} Variations}
\subsection{Concretess Prediction Model}
\label{app:predict_concreteness}
In Table \ref{table: cem ablation}, we report the results obtained by the {\sc CEM} model using  predicted concreteness values (instead of gold standard ones). We predict these values by training the model of  \citet{charbonnier2019predicting} using the concreteness scores of 40k words (all parts-of-speech) in the \citet{brysbaert2014concreteness} dataset. 
We exclude 425 adjectives that are found in the 
{\sc Feature Norms},  {\sc Concept Properties}, and {\sc Memory Colors} datasets.\footnote{In total, the three datasets contain 487 distinct properties (adjectives).}
The concreteness prediction model uses FastText  embeddings \cite{mikolov2018advances} enhanced with POS and suffix features. We evaluate the model on the 425  adjectives that were left out during training and for which we have ground truth scores. The Spearman correlation between the predicted and gold scores is 0.76, showing that our automatically predicted scores can be safely used in our ensemble model instead of the gold standard ones. 

\subsection{CEM Weight Selection} 
\label{app:cem-ablation}

We also experiment with different ways for generating scores and combining the property ranks proposed by the models. (a) \textbf{CEM-pred}:We generate a concreteness score using the model of \citet{charbonnier2019predicting} and FastText embeddings  \cite{bojanowski-etal-2017-enriching}. We train the model on the 40k concreteness dataset \cite{brysbaert2014concreteness}, excluding the 425 adjectives found in our evaluation datasets.
The model obtains a high Spearman $\rho$ correlation of 0.76 against the ground truth scores of the adjectives in our test sets, showing that automatically predicted scores are a good alternative to manually defined ones.
(b)~\textbf{CEM-random}: We randomly generate a score for each property and use it to combine the ranks from two models. (c) \textbf{CEM-average}: We use the average of the property ranks; 
(d) \textbf{CEM-high}: We use the maximum rank of the property; 
(e) \textbf{CEM-low}: We use the minimum rank of the property. Table \ref{table: cem ablation} shows the comparison between {\sc CEM-Pred}, {\sc CEM-Gold}  and   models that rely on these alternative weight generation and ensembling methods on {\sc Feature Norms}. {\sc CEM} achieves the highest performance across all metrics, indicating that  concreteness offers a reliable criterion for model ensembling  under unsupervised scenarios.

\section{Qualitative Analysis}
\subsection{Unigram 
Prediction Frequency}
\label{app: prediction_frequency}
In Table \ref{table:predicted_property_ferquency}, we report the mean Google unigram  frequency \cite{brants2009web} for all  properties in the top 5 predictions of each model. We observe that our {\sc CEM} model -- which achieves the best performance among the tested models, as shown in Table \ref{table: MRD and CSLB main} -- often predicts medium-frequency words. This is a desirable property of our model compared to models which would  instead predict  highly frequent or rare words (highly specific or technical terms). 

This is the case for {\sc GPT3} and {\sc CLIP}, which propose rarer  attributes but obtain lower performance than {\sc CEM}. 
It is worth noting that, contrary to {\sc CLIP}, {\sc GPT3} retrieves properties from an open vocabulary. 

Given that Google NGrams frequencies are computed based on text, many common properties might not be reported. For example, {\sc Feature Norms} propose as typical attributes of an ``ambulance'': 
 {\it loud, white, fast, red, large, orange}. 
The frequency of the corresponding property-noun bigrams (e.g., {\it loud ambulance}, {\it white ambulance}) 
are: 0, 687, 50, 193, 283, and 0. Meanwhile, the bigrams formed with less typical properties (e.g., {\it old, efficient, modern}, and {\it independent}) have a higher frequency (1725, 294, 314, and 457). While language models rely on text and, thus, suffer from reporting bias, vision-based models can retrieve properties that are rarely stated in the text.

\begin{table*}[!ht]
\centering
\scalebox{.9}{
\begin{tabular}{c|c|c|c|c} 
\Xhline{3\arrayrulewidth}
\multirow{2}{*}{{\bf Model}}  &\multicolumn{2}{c|}{
{\sc Concept Properties}{\tt-test}} & \multicolumn{2}{c}{\sc
Feature Norms} \\ \cline{2-5}
&  Unigram Freq. $\downarrow$ & Bigram  Freq. $\downarrow$  & Unigram  Freq. $\downarrow$  & Bigram  Freq. $\downarrow$  \\ \hline
{\sc BERT} & 53M & 11.6K & 55M & 7.6K \\ 
{\sc RoBERTa} & 50M  & 6.8K  &  53M & 6K\\ 
{\sc GPT-2} & 96M  & 10.3K &  78M & 6.4K \\ 
{\sc GPT-3} & 24M  & 6.5K &  25M & 2.8K \\ 
{\sc ViLT} & 50M & 6.2K &  40M & 3.8K \\ 
{\sc CLIP} & 11M  & 5.3K  &  18M&  2.2K \\ 
{\sc CEM-Gold} & 32M  & 7.4K &  33M &  4.1K \\
{\sc CEM-Pred} & 34M  & 7.1K &  31M &  6.1K \\ \Xhline{3\arrayrulewidth}
\end{tabular}}
\caption{Mean Google unigram and bigram frequency for the top-5 predictions by each model. We observe that {\sc CEM} produces rarer words than most other models (excluding {\sc GPT-3} and {\sc CLIP}) while maintaining high performance.}
\label{table:predicted_property_ferquency}
\end{table*}

\subsection{Prototypical Property Retrieval}
\label{app: prototypical}

We carry out an additional experiment aimed at estimating the performance of the models on prototypical vs. non-prototypical properties. Prototypical properties are the ones that apply to most of the objects in the class denoted by the noun (e.g., {\it red strawberries}); in contrast, non-prototypical properties describe attributes of a smaller subset of the objects denoted by the noun (e.g., {\it delicious strawberry}). We make the assumption that prototypical properties are common and, often, visual or perceptual; we expect them to be more rarely stated in texts and, hence, harder to retrieve using language models than using  images.

We use the split of the {\sc Feature Norms} dataset performed by \citet{apidianaki-gari-soler-2021-dolphins} into prototypical and non-prototypical properties, based on the quantifier annotations found in the \citet{herbelot2015building} dataset.\footnote{Three native English speakers were asked to rate properties in {\sc Feature Norms} based on how often they describe a noun, by choosing a label among {\tt [NO, FEW, SOME, MOST, ALL]}.} The first split ({\tt Prototypical}) contains 785 prototypical adjective-noun pairs (for 386 nouns) 
annotated with at least two {\sc all} labels, or with a combination of {\sc all} and {\sc most} (\textit{healthy banana} 
$\rightarrow$ [{\sc all-all-all]}). The second set  ({\tt Non-Prototypical}) contains 807 adjective-noun  pairs (for 509 nouns) with adjectives in the ground truth that are not included in the {\tt Prototypical} set. In Table \ref{table: MRD prototypical}, we report the performance of each model in retrieving these properties. 
In the {\tt ALL, MOST} column we consider properties that have at least 2 {\tt ALL} annotations, with the combination of a {\tt MOST} annotation, and in the {\tt SOME} column, we consider all properties that do not contain {\tt NO} and {\tt FEW} annotations, and have at least one {\tt SOME} annotation. 
The results confirm our intuition that non-prototypical properties are more frequently mentioned in the text. This is reflected in the score of the {\sc Google NGram} baseline for these properties. For prototypical properties, our {\sc CEM }model  outperforms all other models. 

\subsection{Same Top-K Predictions by Different Nouns}
\label{app:duplicates}

\begin{figure*}[!t]
\centering
    \includegraphics[width=\linewidth]{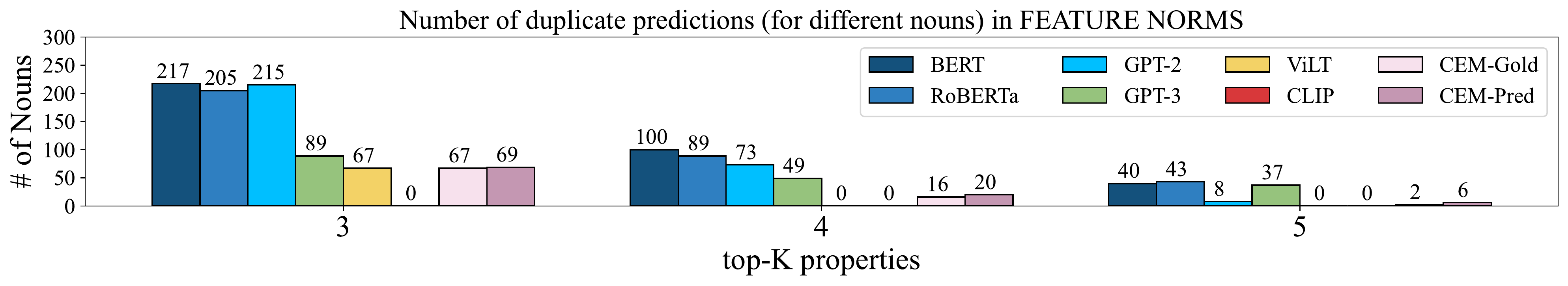}
    \includegraphics[width=\linewidth]{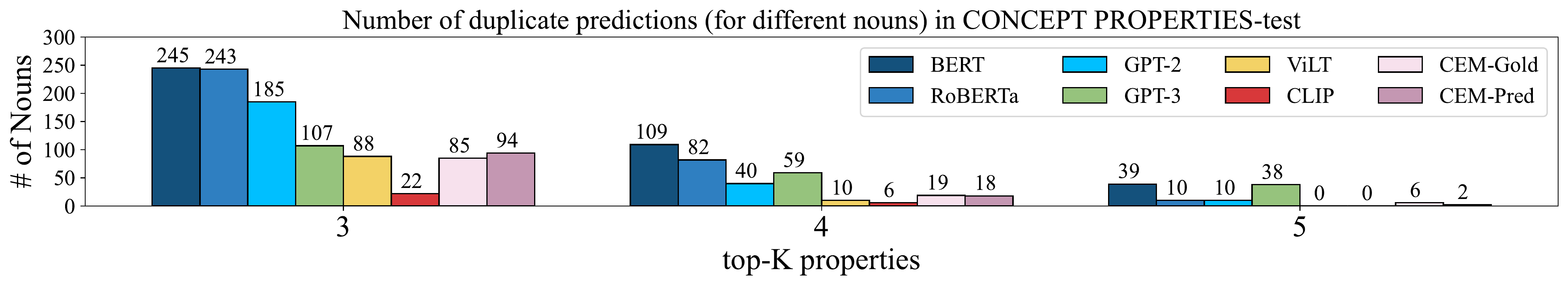}
    
    \caption{Number of nouns in the {\sc Feature Norms} and {\sc Concept Properties}-{\tt test}   datasets for which a model proposed the same top-K  properties (where K = (3,4,5)) in the same order.}
    \label{fig:duplicate_predictions_all}
\end{figure*}

Figure \ref{fig:duplicate_predictions_all} shows the number of nouns in the {\sc Feature Norms} and {\sc Concept Properties}-{\tt test} datasets for which a  model made the exact same top-K predictions. We observe that LMs consistently repeat the same properties for different nouns, while MLMs exhibit a higher variation in their predictions.

\subsection{Multi-piece Performance}
\label{app: multitok_performance}
\begin{table*}[!ht]
\centering
 \resizebox{.8\linewidth}{!}{%
\begin{tabular}{c|cccccc}
\Xhline{3\arrayrulewidth}
\multirow{2}{*}{{\bf Model}}  & \multicolumn{6}{c}{{\sc Feature Norms}}  \\ 
 & {\bf \# Multi-piece Properties} & {\bf Acc@5}       & {\bf Acc@10}  & {\bf R@5}       & {\bf R@10}    & {\bf MRR}      \\ \hline
{\sc BERT-large} & 106 & 0.0 & 0.0 & 0.0 & 0.0 & 0.009  \\
{\sc RoBERTa-large} & 590 & 23.77 & 32.02 & 22.64 & 32.27 & 0.182  \\
{\sc GPT2-large}  & 12 & 0.0 & 0.0 & 0.0 & 0.0 & 0.018 \\
{\sc GPT3-davinci} & 0 & - & - & - & - & -    \\ \hline
{\sc ViLT } & 106 & 1.57 & 2.55 & 7.51 & 13.0 & 0.060   \\
{\sc CLIP-ViT/L14 } & 45 & 4.72 & 5.50 & 55.95 & 66.67 & 0.401  \\
{\sc CEM-Gold (Ours)}  & 590/45 & \textbf{36.54}/1.2 &\textbf{43.81}/3.14 & \textbf{37.65}/13.10 & \textbf{49.59}/35.71 & \textbf{0.245}/0.124 \\
{\sc CEM-Pred (Ours)}  & 590/45 & {32.22}/1.77 &{41.85}/3.73 & {33.7}/20.24 & {46.76}/42.86 & {0.165}/0.122 \\\Xhline{3\arrayrulewidth}
\end{tabular}
 }
\bigskip
 \resizebox{.8\linewidth}{!}{%
\begin{tabular}{c|ccccccc}
\Xhline{3\arrayrulewidth}
\multirow{2}{*}{{\bf Model}}  & \multicolumn{6}{c}{{\sc Concept Properties}{\tt-test}} \\ 
   & {\bf \# Multi-piece Properties}    & {\bf Acc@5}       & {\bf Acc@10} & {\bf R@5}       & {\bf R@10}      & {\bf MRR}       \\ \hline
{\sc BERT-large} & 429 & 0.0 &0.33 & 0.0 & 0.59 & 0.006 \\
{\sc RoBERTa-large} &  1939 & 45.42 & 59.56 & 19.12 & 27.65 & 0.120  \\
{\sc GPT2-large}  & 60 & 0.0 & 0.0 & 0.0 & 0.0 & 0.010 \\
{\sc GPT3-davinci} & 27 & 0.33 & 0.50 & 7.41 & 11.11 & -   \\ \hline
{\sc ViLT}  &  429 & 1.66 & 3.99 & 2.43 & 5.77 & 0.029   \\
{\sc CLIP-ViT/L14}  &  300 & 16.47 & 20.13 & 39.12 & 49.12 & 0.029 \\
{\sc CEM-Gold (Ours)} & 1939/300 & {54.58}/6.49 &{68.39}/9.65 & {26.24}/13.03  & {38.92}/20.96 & {0.161}/0.095 \\
{\sc CEM-Pred (Ours)} & 1939/300 & \textbf{56.99}/5.63 &\textbf{69.87}/9.62 & \textbf{27.31}/12.12  & \textbf{39.35}/21.05 & \textbf{0.165}/0.078 \\\Xhline{3\arrayrulewidth}
\end{tabular}
}
\caption{Performance on multi-piece properties by each model. The highests scores are highlighted in 
{\bf boldface}. 
CEM uses two different tokenizers RoBERTa/CLIP. Hence, we report results for both separated by a backslash (/).} 
\label{tab:multitok_performance}
\end{table*}

Each model splits words into a different number of word pieces. 
Table \ref{tab:multitok_performance} shows the number of multi-piece properties for each model, and its performance on these properties. We observe that all models perform worse than average (refer to Table \ref{table: MRD and CSLB main} for the average performance) on the multi-piece properties, however, {\sc CEM} has the smallest reduction in performance compared to the average values. This could be because {\sc CEM} relies on information from two models with different tokenizers.

\subsection{Qualitative Examples}
\label{app: more_qual}
\vspace{-.15cm}
Table \ref{table:qualitative_results_appendix} contains more 
examples of the top-5 predictions for nouns in {\sc Concept Properties}-{\tt test} and {\sc Feature Norms}. 
\begin{table*}[!ht]
\centering
\scalebox{0.7}{
\begin{tabular}{cccc}
\Xhline{3\arrayrulewidth}
\textbf{Noun}                & \textbf{Image}     & \textbf{Model}   & \textbf{Top-5 Properties}  \\ \hline
                   
\multirow{4}{*}{wand} & \multirow{4}{*}{\cincludegraphics[height=1.8cm]{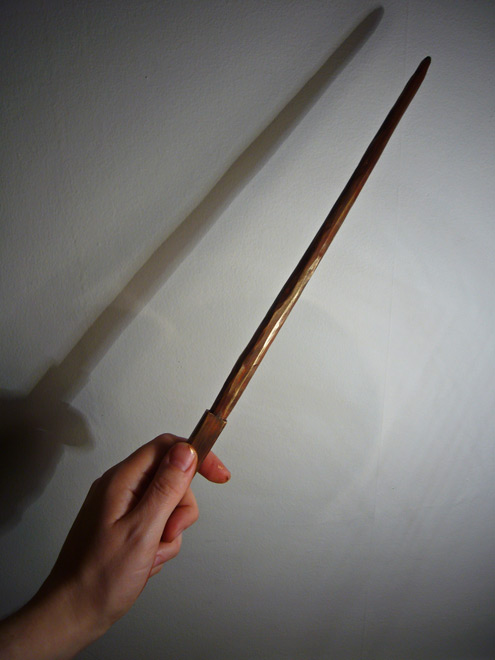}}& {\sc RBTa} & necessary,useful,unnecessary,white,small \\ 
& & {\sc CLIP} & magical,magic,cunning,fizzy,extendable \\ 
& & {\sc GPT-3}  & long,thin,flexible,smooth,light \\ 
& & {\sc CEM-Gold} & magical,long,brown,magic,adjustable \\ 
& & {\sc CEM-Pred} & magical,brown,magic,long,golden \\ \hline

\multirow{4}{*}{horse} & \multirow{4}{*}{\cincludegraphics[width=1.8cm]{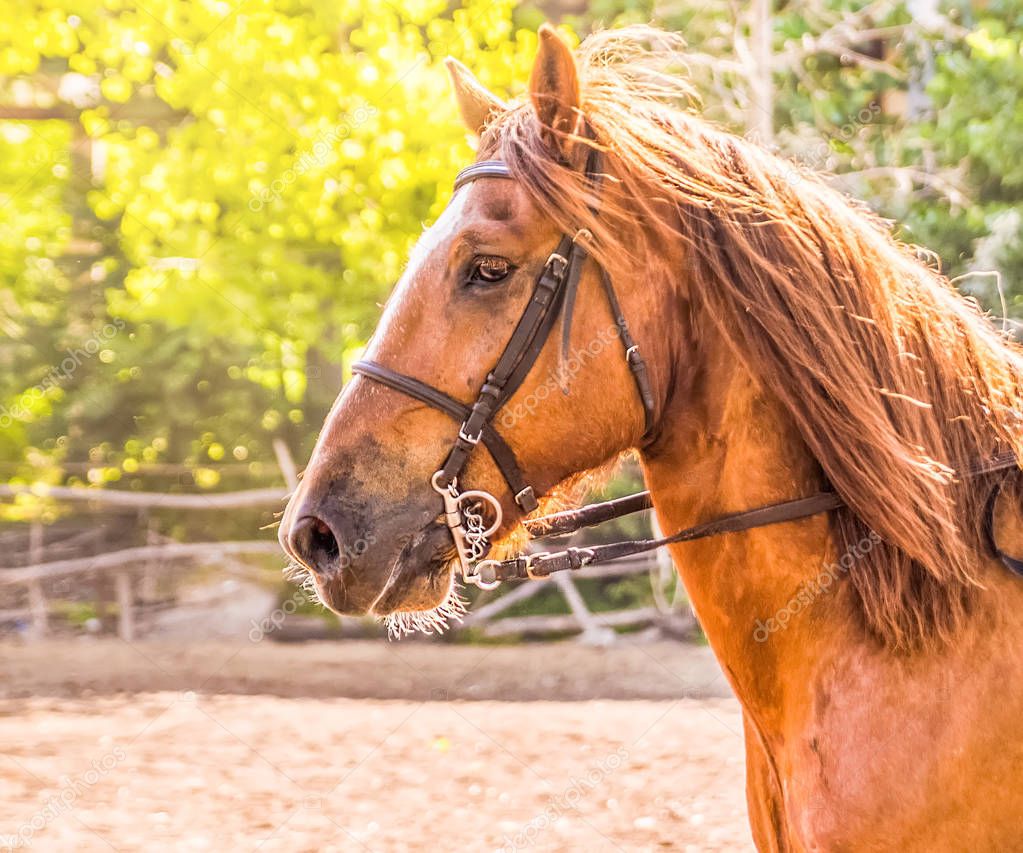}} & {\sc RBTa} & healthy,white,black,stable,friendly \\ 
& & {\sc CLIP} & stable,majestic,fair,free,wild \\ 
& & {\sc GPT-3} & strong,fast,powerful,muscular,big \\  
& & {\sc CEM-Gold} & stable,friendly,free,wild,healthy \\ 
& & {\sc CEM-Pred} &  stable,friendly,free,wild,healthy \\ \hline

\multirow{4}{*}{raven} & \multirow{4}{*}{\cincludegraphics[height=1.8cm]{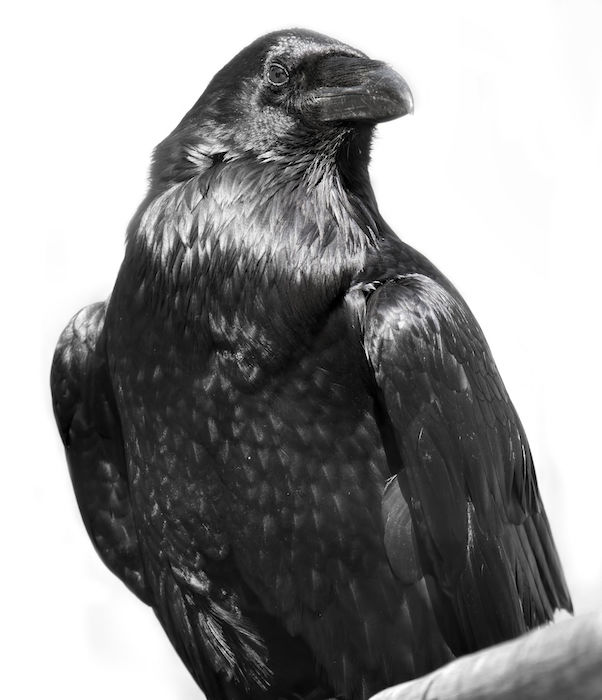}} & {\sc RBTa} & black,white,harmless,aggressive,solitary \\ 
& & {\sc CLIP} & unlucky,nocturnal,dark,solitary,cunning \\ 
& & {\sc GPT-3} &  black,glossy,sleek,shiny,intelligent \\ 
& & {\sc CEM-Gold} & solitary,black,dark,harmless,rare \\ 
& & {\sc CEM-Pred} & solitary,black,dark,harmless,rare \\ \hline

\multirow{4}{*}{surfboard} & \multirow{4}{*}{\cincludegraphics[height=1.8cm]{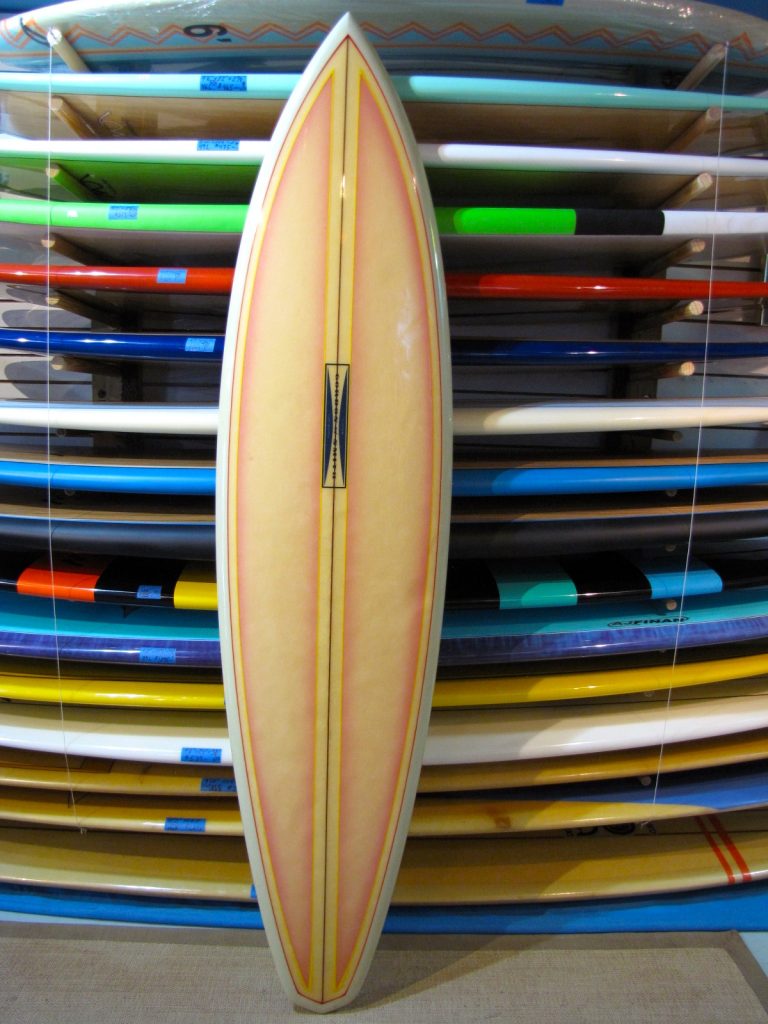}}& {\sc RBTa} & expensive,white,comfortable,small,waterproof \\ 
& & {\sc CLIP} & paddled,overfished,aerodynamic,concave,beachwear \\ 
& & {\sc GPT-3} & hard,smooth,slick,colorful,long \\ 
& & {\sc CEM-Gold} & waterproof,paddled,beachwear,long,cool \\ 
& & {\sc CEM-Pred} & waterproof,long,cheap,cool,durable \\ \hline

\multirow{4}{*}{limousine} & \multirow{4}{*}{\cincludegraphics[width=1.8cm]{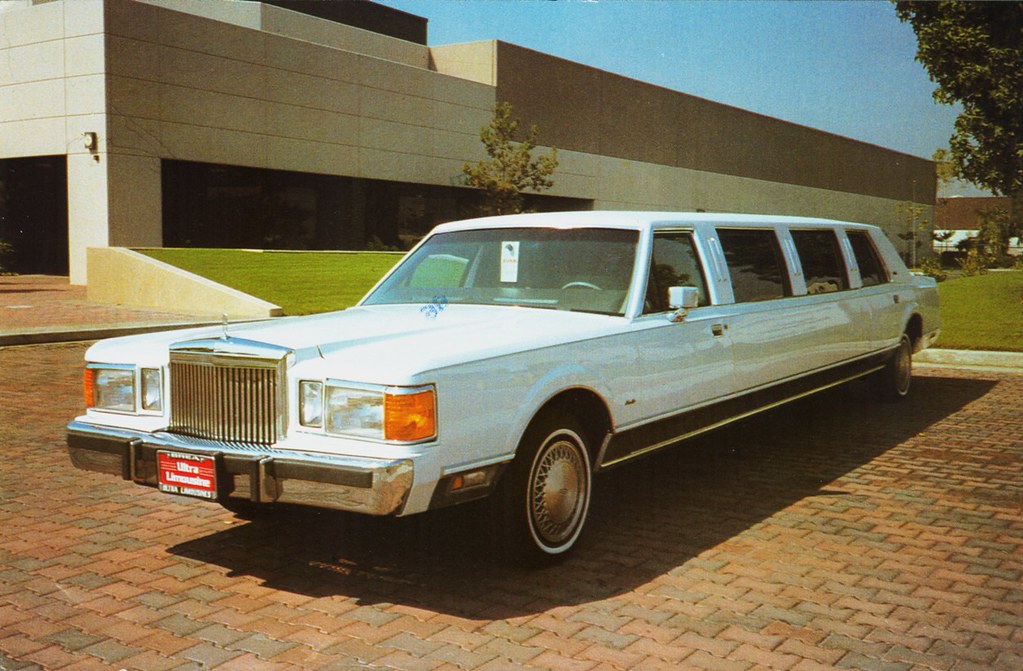}}& {\sc RBTa} & expensive,black,white,empty,large \\ 
& & {\sc CLIP}&  luxurious,decadent,ostentatious,expensive,showy \\ 
& & {\sc GPT-3} & long,sleek,spacious,luxurious,comfortable \\
& & {\sc CEM-Gold} & expensive,luxurious,large,long,comfortable \\ 
& & {\sc CEM-Pred} & expensive,luxurious,large,long,comfortable  \\ \hline

\multirow{4}{*}{violin} & \multirow{4}{*}{\cincludegraphics[height=1.8cm]{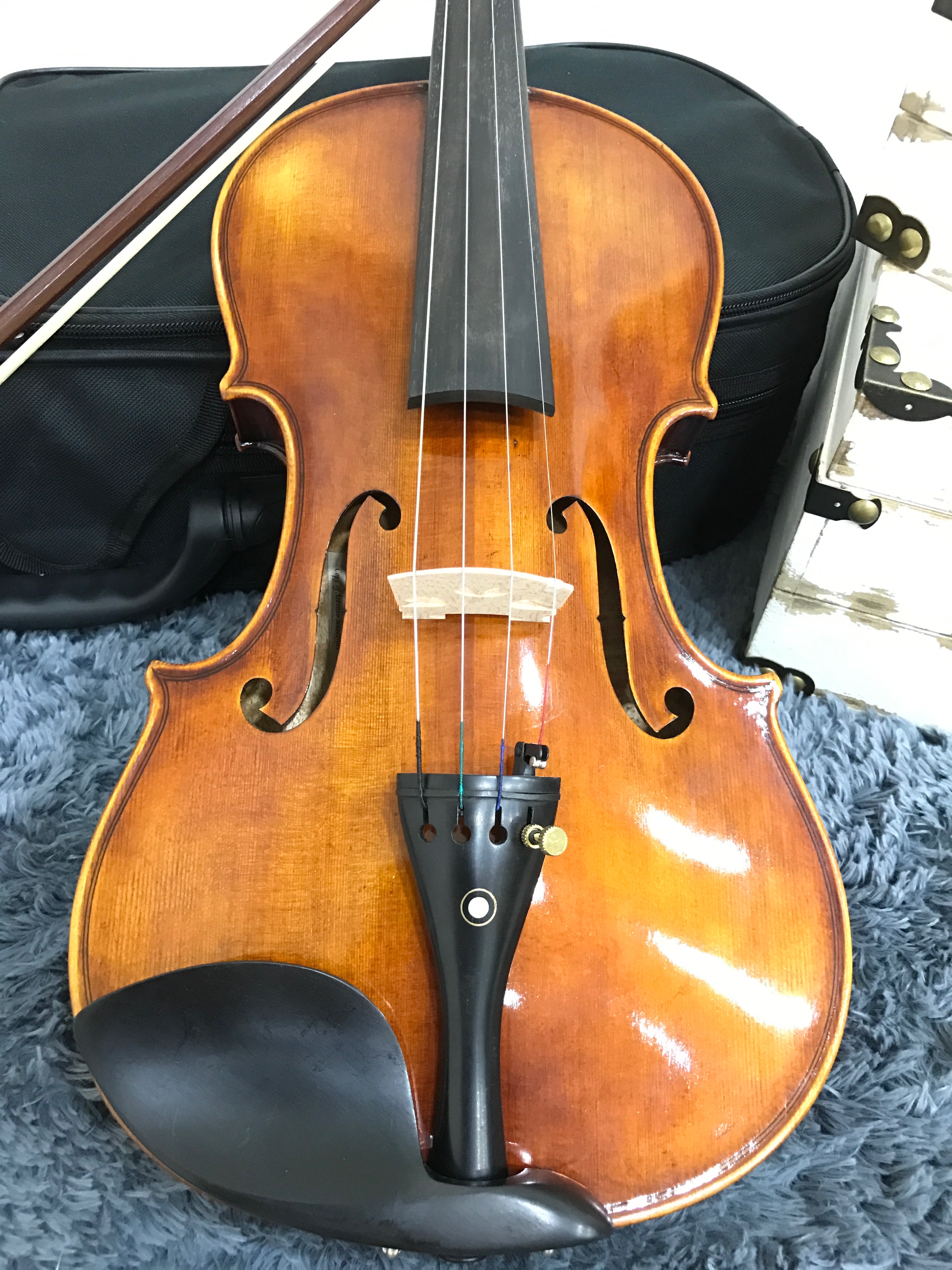}}& {\sc RBTa} & expensive,white,small,black,electric \\ 
& & {\sc CLIP} & acoustic,fiddly,strummed,traditional,rhythmic \\ 
& & {\sc GPT-3}  & wooden,long,thin,stringed,musical \\ 
& & {\sc CEM-Gold} & acoustic,fiddly,small,cheap,unique \\ 
& & {\sc CEM-Pred} & acoustic,small,unique,cheap,brown \\ \hline

\multirow{4}{*}{barn} & \multirow{4}{*}{\cincludegraphics[width=1.8cm]{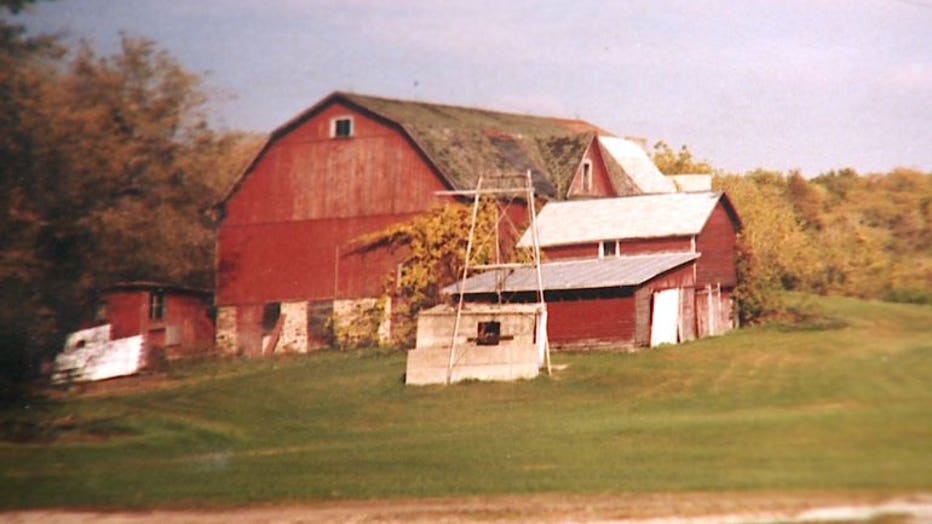}}& {\sc RBTa} & large,white,small,red,common \\ 
& & {\sc CLIP} & old-fashioned,run-down,harvested,red,old \\ 
& & {\sc GPT-3} & old,large,red,wooden,rusty \\ 
& & {\sc CEM-Gold} & red,large,old,spacious,portable \\ 
& & {\sc CEM-Pred} & red,old,spacious,large,rectangular \\ \hline

\multirow{4}{*}{oak} & \multirow{4}{*}{\cincludegraphics[height=1.8cm]{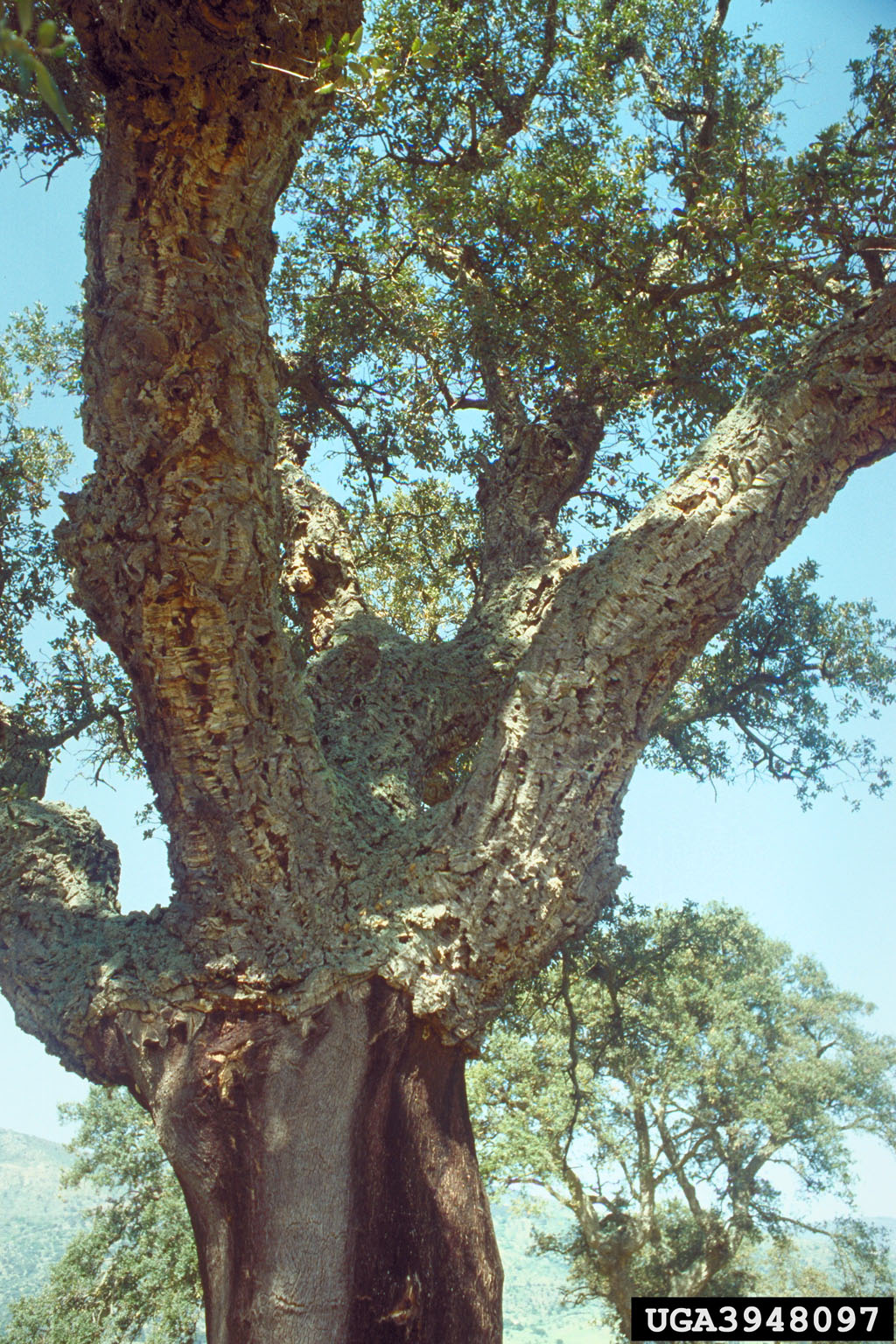}}& {\sc RBTa} & healthy,white,green,tall,harvested \\ 
& & {\sc CLIP} & green,sticky,edible,large,harvested \\ 
& & {\sc GPT-3} & strong,sturdy,hard,dense,heavy \\
& & {\sc CEM-Gold} & green,harvested,large,edible,brown \\ 
& & {\sc CEM-Pred} & green,harvested,large,edible,brown \\ \hline

\multirow{4}{*}{radish} & \multirow{4}{*}{\cincludegraphics[height=1.8cm]{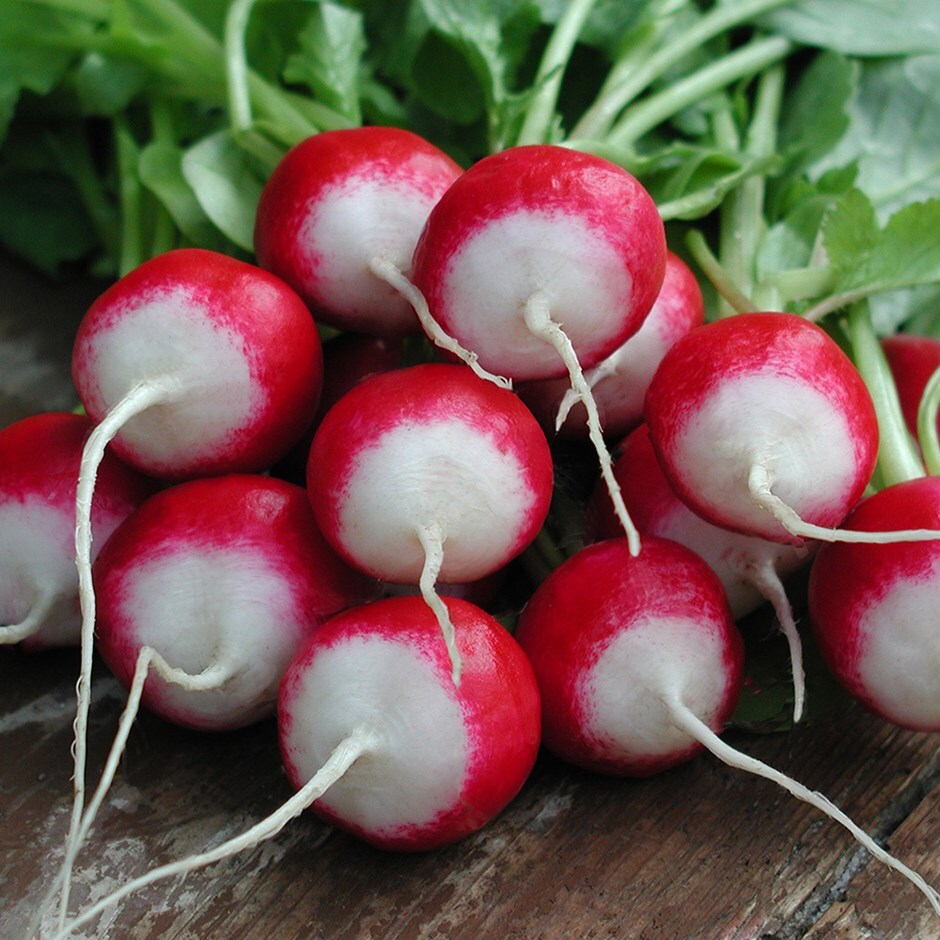}}& {\sc RBTa} & edible,poisonous,delicious,white,small \\ 
& & {\sc CLIP} & edible,nutritious,healthy,harvested,young \\ 
& & {\sc GPT-3}  & crunchy,peppery,spicy,earthy,pungent \\ 
& & {\sc CEM-Gold} & edible,healthy,harvested,delicious,white \\ 
& & {\sc CEM-Pred} & edible,white,harvested,healthy,delicious \\ \hline

\multirow{4}{*}{toilet} & \multirow{4}{*}{\cincludegraphics[height=1.8cm]{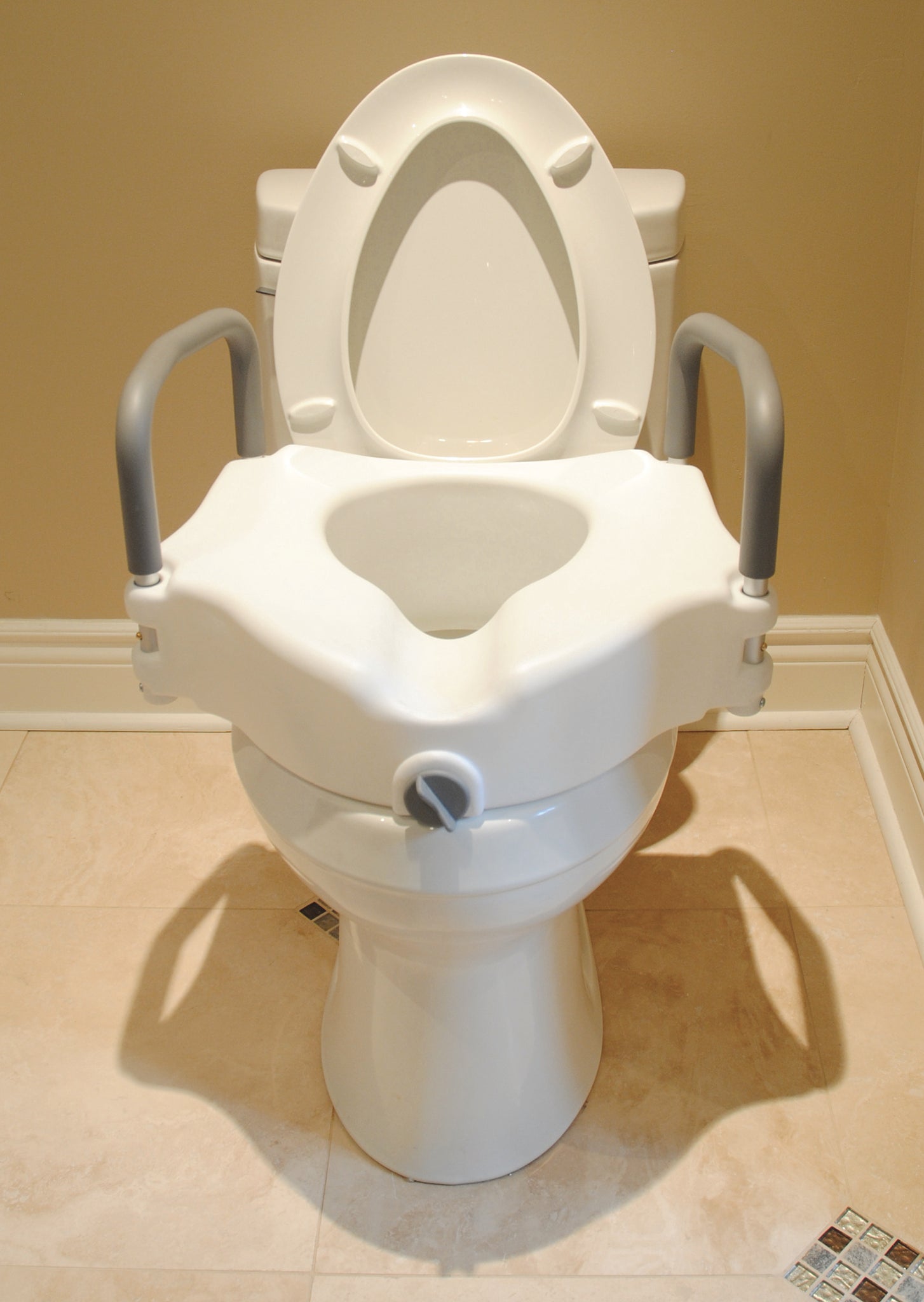}} & {\sc RBTa} & portable,dirty,open,common,small \\ 
& & {\sc CLIP} & sinkable,brown,emptied,round,short \\ 
& & {\sc GPT-3} &  dirty,smelly,clogged,rusty,filthy \\
& & {\sc CEM-Gold} &large,white,small,brown,sinkable \\ 
& & {\sc CEM-Pred} &portable,white,uncomfortable,waterproof,dirty\\ \hline

\multirow{4}{*}{bluejay} & \multirow{4}{*}{\cincludegraphics[height=1.8cm]{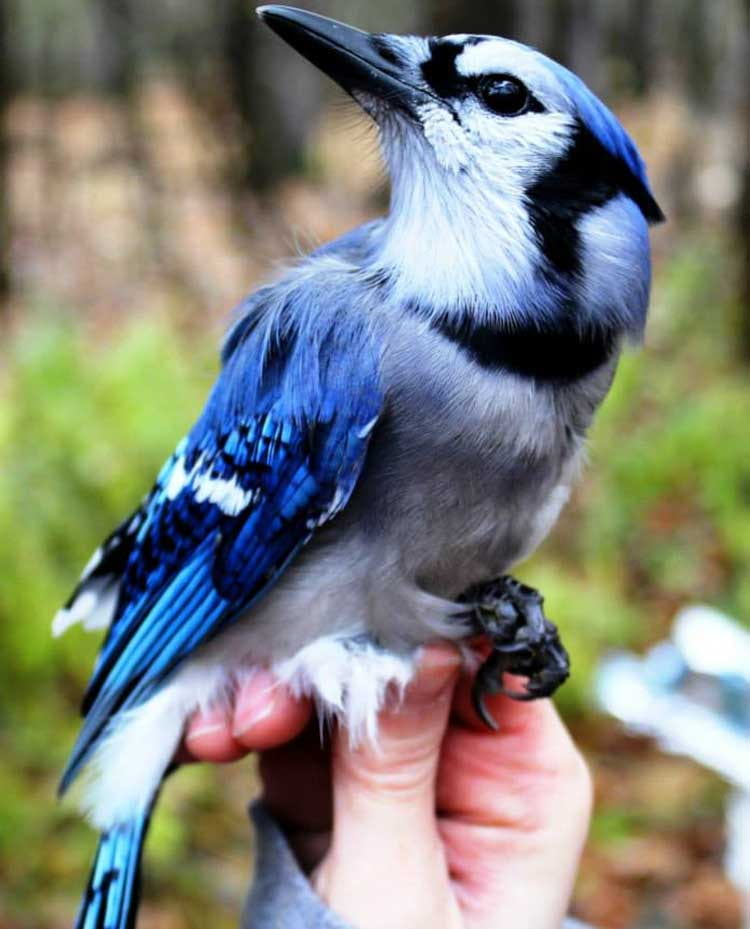}}& {\sc RBTa} & male,black,small,healthy,white \\ 
& & {\sc CLIP} & blue,gentle,crunchy,wild,friendly \\ 
& & {\sc GPT-3} & blue,small,blue,blue,blue \\ 
& & {\sc CEM-Gold} & blue,friendly,edible,small,endangered \\ 
& & {\sc CEM-Pred} & blue,endangered,small,friendly,wild \\ \hline

\multirow{4}{*}{donkey} & \multirow{4}{*}{\cincludegraphics[height=1.8cm]{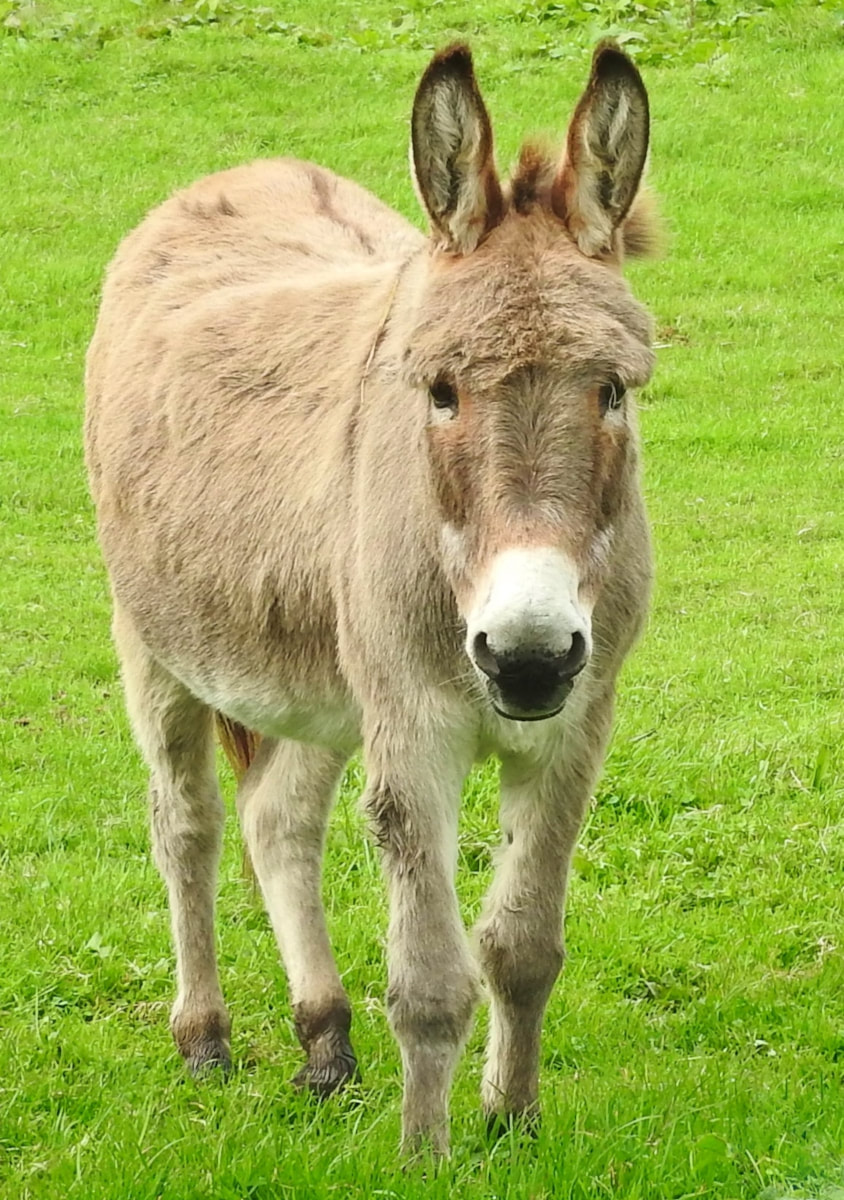}}& {\sc RBTa} &black,male,white,healthy,brown \\ 
& & {\sc CLIP}& humorous,annoying,short,brown,darned \\ 
& & {\sc GPT-3} & stubborn,strong,sure-footed,intelligent,social \\
& & {\sc CEM-Gold} &brown,grey,large,short,slow \\ 
& & {\sc CEM-Pred} & brown,large,slow,hairy,friendly \\ \hline

\end{tabular}
}
\caption{Random sample of Top-5 properties proposed by different models for nouns in the {\sc Feature Norms} and {\sc Concept Properties}{\tt-test} dataset.}
\label{table:qualitative_results_appendix}
\end{table*}

\end{document}